\documentclass{article}
\PassOptionsToPackage{numbers}{natbib}
\usepackage[preprint]{neurips_2026}

\usepackage[utf8]{inputenc}
\usepackage[T1]{fontenc}
\usepackage{hyperref}
\usepackage{url}
\usepackage{booktabs}
\usepackage{amsfonts}
\usepackage{nicefrac}
\usepackage{microtype}
\usepackage{xcolor}

\usepackage{amsmath}
\usepackage{amssymb}
\usepackage{mathtools}
\usepackage{graphicx}
\usepackage{multirow}
\usepackage{array}
\usepackage{float}  
\usepackage{algorithm}
\usepackage{algpseudocode}

\newcommand{\NARC}{\textsc{NARC}}
\newcommand{\eps}{\epsilon}
\newcommand{\epsstar}{\eps^{\star}}
\newcommand{\xtstar}{x_T^{\star}}
\newcommand{\hateps}{\hat{\eps}}

\title{Compression Asymmetry and Trajectory Binding in Noise-Anchored Diffusion Inversion}

\author{%
  Yongseong Park \\
  Department of EECS \\
  DGIST \\
  Daegu, Republic of Korea \\
  \texttt{reo91004@dgist.ac.kr} \\
  \And
  Joeun Kim \\
  Department of AI \\
  DGIST \\
  Daegu, Republic of Korea \\
  \texttt{jowithu@dgist.ac.kr} \\
  \And
  HoEun Kim \\
  Department of AI \\
  DGIST \\
  Daegu, Republic of Korea \\
  \texttt{hoeun.kim@dgist.ac.kr} \\
  \AND
  Young-Sik Kim\thanks{Corresponding author.} \\
  Department of EECS, Department of AI \\
  DGIST \\
  Daegu, Republic of Korea \\
  \texttt{ysk@dgist.ac.kr} \\
}

\begin{document}

\maketitle

\begin{abstract}
Real-image diffusion inversion is governed by a tight quality--cost
trade-off, with costs incurred in computation, storage, or per-image
optimization. We study this trade-off through the forward Gaussian
noise anchor that defines a diffusion trajectory and isolate two
mechanisms behind effective stored-noise inversion. First, diffusion
noise exhibits an \emph{element-wise compression asymmetry}: int8
full-dimensional anchors preserve reconstruction, whereas
low-dimensional subspace summaries are much less reliable, often
collapsing even at comparable or smaller payloads; the element-wise
over subspace ordering persists across five stored-noise inversion
methods. Second, inversion is
\emph{trajectory-bound and score-prior coupled}: the matched forward
anchor and a trained score network are both necessary, arguing
against a purely algebraic-identity explanation. Together, these
findings specify what to store and how to use it. They lead to
\textbf{Noise-Anchored Reverse Correction (\NARC{})}, a training-free
inversion primitive that stores a single int8 latent anchor and
reuses it with a fixed, noise-level-dependent anchor-weight schedule:
strong anchoring when the reverse trajectory is noise-dominated, then
relaxed anchoring as image detail emerges. On PIE-Bench++ with
Stable Diffusion~1.5, \NARC{} outperforms five modern non-exact
baselines and improves PSNR by $+3.24$\,dB over PnP DirectInv while
using about $400\times$ less inversion storage than PnP DirectInv.
The compression asymmetry, anchor specificity, and editing plug-in
also transfer to SDXL~$1024^2$.
\end{abstract}

\section{Introduction}
\label{sec:intro}

Inverting a real image into the latent space of a pretrained
text-to-image diffusion model~\citep{rombach2022sd} is a prerequisite
for reconstruction and editing, and is governed by a tight
cost--quality trade-off. Existing approaches pay this cost in
different forms: deterministic forward DDIM unrolling
\citep{song2020ddim}, per-image embedding optimization
\citep{mokady2022nulltext}, per-step iterative refinement
\citep{garibi2024renoise}, dual-condition fixed-point optimization
\citep{li2025dci}, dual coupled streams enforcing algebraic
invertibility \citep{wallace2022edict}, or caching all
forward-step noise residuals
\citep{huberman2024editfriendly,ju2023pnp}. All sit on a single
``quality $\Rightarrow$ NFE / storage / per-image optimization'' axis.

\paragraph{This paper.}
We approach inversion not by adding another point on this axis, but
by asking what the closed-form forward marginal already indicates.
The DDPM forward distribution \citep{ho2020ddpm} is
\begin{equation}
x_t = \sqrt{\bar\alpha_t}\,x_0+\sqrt{1{-}\bar\alpha_t}\,\eps,
\qquad \eps\sim\mathcal{N}(0,I),
\label{eq:forward}
\end{equation}
Throughout the paper, $x_t$ denotes the state evolved by the
diffusion sampler. For latent diffusion models such as SD~1.5 and
SDXL, this state is the VAE latent often written as $z_t$; for the
CIFAR pixel DDPM validation checks, it is the pixel-space diffusion
state. All storage numbers are computed in this diffusion-domain
state space.
With this convention, sampling $\epsstar$ once and forming
$\xtstar=\sqrt{\bar\alpha_T}\,x_0+\sqrt{1{-}\bar\alpha_T}\,\epsstar$
makes $\epsstar$ a \emph{timestep-independent anchor} of that
trajectory. Note $\epsstar$ alone does not encode $x_0$; the
information lives in the relation between $\xtstar$ and $\epsstar$.
Because the $\eps$-prediction model is trained to recover the noise
component at every $t$, predictions $\hateps_t$ remain close to
$\epsstar$ when reverse trajectories stay near the forward one.
This suggests the one-line correction summarized in Fig.~1:
at every reverse step, blend the model's prediction with the cached anchor,
\begin{equation}
\eps_t^{\text{corr}} = (1-\lambda_t)\,\hateps_t+\lambda_t\,\epsstar,
\qquad \lambda_t\in[0,1].
\label{eq:blend}
\end{equation}

\begin{figure}[t]
\centering
\includegraphics[width=\textwidth]{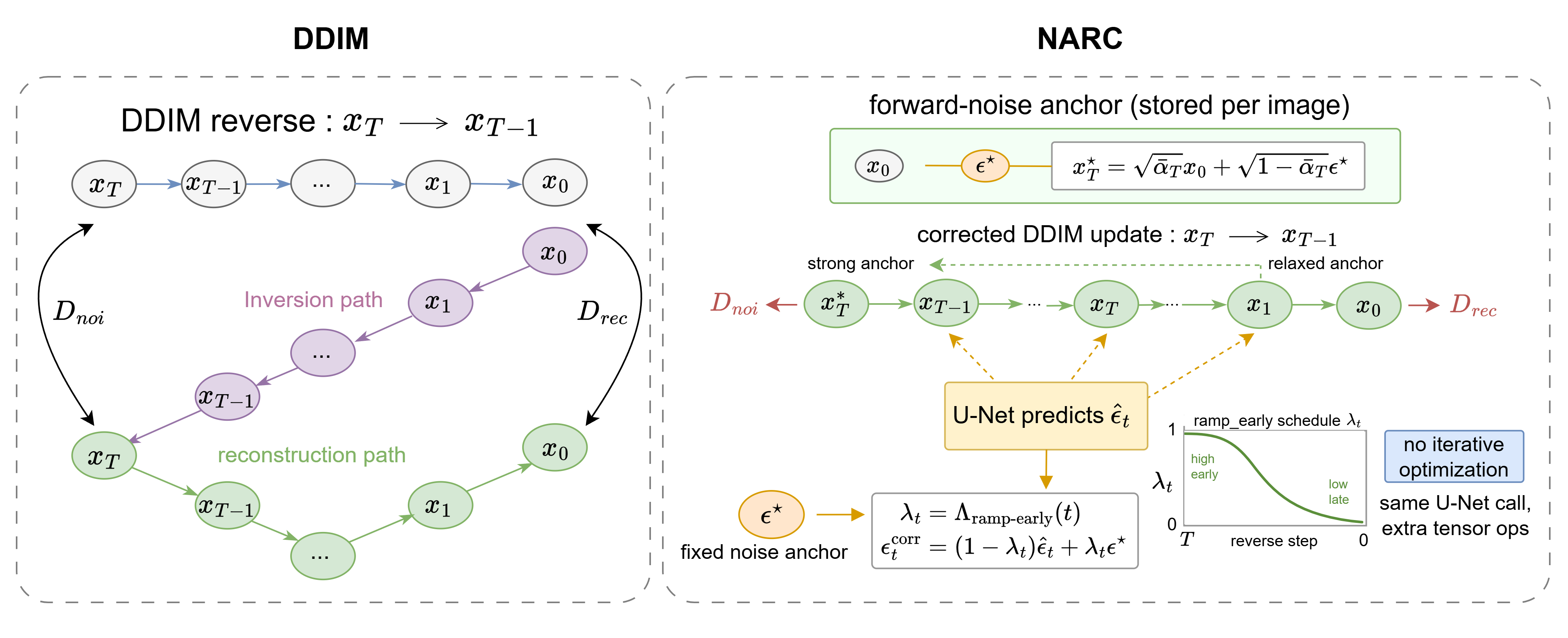}
\vspace{-20pt}
\caption{\textbf{NARC overview.} A forward-noise anchor
$\epsstar$ is sampled once per image, stored as the per-image cache,
and reused during deterministic reverse sampling by blending the
model prediction with the anchor under the scheduled weight
$\lambda_t$. All variables follow the diffusion-domain convention
defined in the text.}
\label{fig:overview}
\end{figure}

\paragraph{Two mechanism findings.}
Studying when and why this simple correction works leads to two
empirical findings that, we argue, are the actual contribution of
this paper.

\textbf{Finding~1: Element-wise vs.\ subspace compression asymmetry.}
Standard intuition from image processing suggests that subspace
projections (DCT low band, low-rank approximation) preserve
structured information efficiently, while element-wise quantization
is wasteful. We find the opposite holds for diffusion noise anchors.
int8 4$\times$ element-wise quantization of $\epsstar$ is
statistically near-lossless in cache-accounted SD/SDXL audits,
while low-dimensional DCT, random-projection, spatial-mask, and
block-average summaries at comparable or smaller payloads are much
less reliable: on SD they lose PSNR and LPIPS, and on SDXL they can
raise PSNR through smoothing while still worsening LPIPS.
Testing the same six compression schemes on five stored-noise
inversion methods reveals a 3-tier magnitude gradient
(\emph{Strong} / \emph{transition} / \emph{boundary}), governed by
two conditions: per-dimension cache magnitude must exceed the
quantization step, and cache information must be isotropically
dispersed across ambient space. The asymmetry is therefore a
paradigm-wide property of the noise latent space, not a
NARC-specific artifact (\S\ref{sec:mech:f1}).

\textbf{Finding~2: Trajectory binding and score-prior coupling.}
Three independent measurements jointly rule out an ``algebraic
identity'' explanation. Here and below, anchor specificity refers to
the matched $(\xtstar,\epsstar)$ trajectory pair, not to $\epsstar$
alone as a source code. The trajectory-matched forward anchor
reconstructs, while random / mismatched-image / shuffled / sign-flipped
correction anchors collapse with a $+18.17$--$18.23$\,dB paired gap
on the main schedule ---
($+10.98$--$11.08$\,dB at the conservative reference
$\lambda{=}0.7$). Replacing the trained UNet with a random-weight
UNet collapses reconstruction by $-11.75$\,dB at the main schedule
($-10.18$\,dB at fixed $\lambda{=}0.7$). And classifier-free guidance (cfg) 
sensitivity is unimodal in $\lambda$: the main schedule sits just below the fixed
$\lambda{=}0.9$ comparator in mean anchor weight, yet retains
positive cfg sensitivity and trained-model dependence. The procedure
is therefore a \emph{model-coupled anchor correction regime}, not an
algebraic inversion endpoint (\S\ref{sec:mech:f2}).

\paragraph{From mechanism to method.}
\NARC{} (Noise-Anchored Reverse Correction) is the minimal
training-free primitive that operationalises the two findings: a
closed-form forward jump (zero forward UNet calls) followed by
reverse drift correction reusing $\epsstar$ via Eq.~\ref{eq:blend}.
The diffusion-geometric prescription ``high anchor at high noise,
model-driven texture at low noise'' is realized by the
deterministic schedule
$\lambda(\bar\alpha_t)=\lambda_{\max}-(\lambda_{\max}-\lambda_{\min})\,\bar\alpha_t^{\,\gamma}$
with $(\lambda_{\min},\lambda_{\max},\gamma){=}(0.7,0.95,2)$
(\texttt{ramp\_early}, \S\ref{sec:method}). The $\gamma{=}1$
schedule is retained as an ablation, and the conservative fixed
$\lambda{=}0.7$ remains the reference operating point for the
Finding~2 mechanism analyses.

\vspace{-5pt}
\paragraph{Application: a 16\,KiB extreme-compression stored-noise point.}
On PIE-Bench++ SD~1.5 cfg=7.5 ($N{=}140$, paired Wilcoxon by
\texttt{image\_id}), \NARC{} (\texttt{ramp\_early} $\gamma{=}2$,
int8 anchor) is paired $+7.77$ to $+9.79$\,dB above DDIM-Inv,
ReNoise, DCI, FreeInv, and taba on PSNR (all
$p\!\leq\!10^{-24}$, the test floor) with proportionally large
LPIPS gains ($\Delta{=}{-}0.185$ to $-0.351$). \NARC{} further
outperforms PnP DirectInv (method within the same paradigm, 250 NFE,
6.4\,MB cache) by
$+3.24$\,dB at $400\times$ less storage, and shrinks the gap to
Edit-friendly DDPM ($-9.10\!\to\!-1.97$\,dB), Null-text
($-8.39\!\to\!-1.16$\,dB), and EDICT ($-9.39\!\to\!-2.25$\,dB)
(\S\ref{sec:exp:main}). Finding~1, the anchor-specificity component
of Finding~2, and the editing plug-in (\S\ref{sec:exp:editing})
transfer to SDXL~$1024^2$. Storage numbers here and below are
cache-accounted for the retained-source editing setting: the source
image or clean diffusion state is available, and the materialized
per-image cache is the additional quantized anchor.

\vspace{-5pt}
\paragraph{Summary of contributions.}
(i) Two mechanism findings about the diffusion noise latent space
that sharpen prior work on noise--image inversion relations and
forward-jump hybrids (\S\ref{sec:mechanism}); (ii) \NARC{}, a one-line training-free
inversion primitive with a deterministic noise-aware schedule
derived from those findings (\S\ref{sec:method});
(iii) a paired-comparison study placing \NARC{} as a 16\,KiB
additional-cache point in the retained-source editing setting within
the stored-noise paradigm
(\S\ref{sec:exp:main}); (iv) an editing
plug-in showing improved background preservation while target
alignment remains close to the no-\NARC{} baseline
(\S\ref{sec:exp:editing}).

\section{Related Work}
\label{sec:related}

DDPM \citep{ho2020ddpm} and DDIM \citep{song2020ddim} provide the
forward/reverse formalism we follow. \emph{DDIM Inversion} reverses
the deterministic update to obtain $x_T^{\text{inv}}$; it is nearly
lossless at cfg=1 but degrades sharply at cfg$>$1.
\emph{Null-text Inversion} \citep{mokady2022nulltext} optimizes the
unconditional embedding per image (NFE $\geq 600$).
\emph{ReNoise} \citep{garibi2024renoise} refines each forward step
with $K$ inner iterations (NFE $T(K{+}1)$).
\emph{DCI} \citep{li2025dci} uses dual-conditional fixed-point
optimization. \emph{EDICT} \citep{wallace2022edict} guarantees
algebraic exact-invertibility via dual coupled streams.
\emph{FreeInv} \citep{bao2025freeinv} augments DDIM-Inv with rotation;
\emph{taba} \citep{staniszewski2026taba} replaces the first $N$
inversion steps with closed-form forward jumps.
Prior work therefore already studies noise--image inversion
relations and forward-jump hybrids; our claims are the
cache-compression asymmetry across stored-noise objects and its use in
a single-anchor scheduled correction primitive.
\emph{Edit-friendly DDPM} \citep{huberman2024editfriendly} caches
$T{+}1$ noise maps for perfect reconstruction (3.2\,MB at $T{=}50$).
\emph{PnP DirectInv} \citep{ju2023pnp} caches $T$ residuals and
introduces the PIE-Bench++ benchmark.

These methods differ not only in accuracy--cost trade-offs, but also in
the object preserved across inversion and reconstruction: optimized
conditioning or fixed-point states, exact coupled sampler states,
trajectory-level noise maps/residuals, or repaired starting-noise
relations. This view motivates our cache question: whether trajectory
information must be stored as a tensor sequence, or can be reduced to
one forward-marginal anchor and re-coupled through the sampler.

This cache-as-representation view separates our goal from post hoc
compression of a finished inversion method. In the methods above, the
inversion mechanism first determines the retained object---an embedding,
coupled state, terminal latent, noise-map trajectory, or residual
sequence---and storage is then a consequence of that choice. We instead
ask which cached quantity is stable under coarse precision, which
summaries destroy trajectory information, and how the retained object
should be coupled back to the score model. Finding~1 tests this question
across stored-noise cache objects rather than only on the proposed
anchor.

\NARC{} is closest to Edit-friendly~\citep{huberman2024editfriendly}
and PnP DirectInv~\citep{ju2023pnp}, but replaces their trajectory of
cached tensors with one sampled $\epsstar$ (16\,KiB int8 for SD~1.5
latents). Finding~1 (\S\ref{sec:mech:f1}) shows that the qualitative
compression behavior is shared; the new question is how much trust to
place, at each noise level, in one forward-marginal anchor.

\section{Two Mechanism Findings}
\label{sec:mechanism}

This section presents the two empirical findings that motivate the
method. We defer setup details (datasets, statistical protocol, NFE
accounting) to \S\ref{sec:exp:setup} so the reader can follow the
findings without interruption; throughout this section, all SD\,1.5
numbers are PIE-Bench++ \texttt{0\_random\_140} ($N{=}140$) and all
$p$-values are paired Wilcoxon signed-rank by \texttt{image\_id}.

\subsection{Finding 1: An element-wise vs.\ subspace compression asymmetry}
\label{sec:mech:f1}
\label{sec:exp:f1}

\paragraph{The asymmetry.}
A natural intuition from image processing is that subspace projections
(DCT low band, low-rank approximation, spatial masks) preserve
structured information efficiently, so they should be the right tool
for compressing a cached anchor. We find the opposite for the
diffusion noise anchor $\epsstar$. Element-wise int8 quantization
($4{\times}$ smaller than fp32 byte accounting and $2{\times}$
smaller than the measured fp16 full-anchor payload; 16\,KiB on
SD~1.5 latents) is statistically near-lossless in a strict
cache-accounted fp32-runtime audit: the stored payload is decoded
once and the same dequantized anchor is used to rebuild the reverse
start and in the correction term. Under this protocol, fp16 and int8
are indistinguishable from the
fp32 full anchor (int8 $\Delta$PSNR $=+0.00008$\,dB, $p{=}0.89$),
and a packed-int4 anchor remains within $0.12$\,dB
(Appendix~\ref{app:f1}). In contrast, DCT, random-projection,
spatial-mask, and block-average summaries at smaller or comparable
payloads lose $0.99$--$1.41$\,dB on PSNR and worsen LPIPS by
$0.156$--$0.180$ on the same SD~1.5 audit.
On SDXL under the main \texttt{ramp\_early} $\gamma{=}2$ schedule,
the direct cache-accounted audit in Tab.~\ref{tab:int8-lossless}
again finds int8 indistinguishable from the full anchor
($\Delta$PSNR $=+0.00006$\,dB, $p{=}0.744$; $\Delta$LPIPS
$=-0.000055$, $p{=}0.271$). The fuller cache-accounted sweep in
Appendix~\ref{app:sdxl} shows that support-reduced summaries are
non-Pareto: several increase PSNR/SSIM through smoothing, but all
worsen LPIPS by $0.042$--$0.060$. Thus the robust finding is not merely
``subspace PSNR collapse'' in every regime; it is that preserving the
full coordinate support of $\epsstar$ under coarse element-wise
precision is the stable operating point.
The 16\,KiB int8 anchor used throughout the rest of the paper is
therefore directly measured for SD~1.5 at $512^2$ resolution. SDXL
experiments below use the standard fp16 runtime/full-anchor baseline
for memory and numerical stability; their storage ratios are still
reported against fp32 byte accounting.

\begin{table}[!htbp]
\centering
\caption{\textbf{int8 vs.\ full-anchor cache-accounted audit:
4-metric paired Wilcoxon by image/sample id, two-sided.}
Each int8 row decodes one stored $\epsstar$ payload and uses it both
to rebuild $\xtstar$ and in the correction term. Differences are
computed against the corresponding full-anchor row. $L_{0.5}$: mechanism setup
($\lambda{=}0.5$, cfg=1; CIFAR uses its canonical hard-threshold
setting, whereas SD uses always blend). $L_{0.7}$: conservative reference
($\lambda{=}0.7$, cfg=7.5, after-CFG). \texttt{ramp\_early}: main
schedule ($\lambda_{\min}{=}0.7$, $\lambda_{\max}{=}0.95$,
$\gamma{=}2$, mean $\bar\lambda{\approx}0.885$, cfg=7.5). All rows
use the decoded payload consistently at both anchor use sites.
CIFAR LPIPS uses the standard
$224{\times}224$ upsampling for the AlexNet feature backbone.}
\label{tab:int8-lossless}
\small
\setlength{\tabcolsep}{4pt}
\resizebox{\textwidth}{!}{%
\begin{tabular}{llrrrr}
\toprule
Setup & $n$ & $\Delta$PSNR (dB, $p$) & $\Delta$SSIM ($p$) & $\Delta$LPIPS ($p$) & $\Delta$MSE ($\!\times\!10^{-3}$, $p$)\\
\midrule
CIFAR pixel DDPM validation, $L_{0.5}$
& $2000$ & $-0.00065$ ($p{=}0.573$)
& $+0.000005$ ($p{=}0.995$)
& $+0.000021$ ($p{=}0.828$)
& $-0.0030$ ($p{=}0.988$)\\
SD latent, $L_{0.5}$
& $140$ & $+0.00008$ ($p{=}0.894$)
& $-0.000007$ ($p{=}0.088$)
& $-0.000041$ ($p{=}0.551$)
& $+0.0005$ ($p{=}0.784$)\\
SD latent, $L_{0.7}$ ref.
& $140$ & $+0.0028$ ($p{=}0.574$)
& $+0.000004$ ($p{=}0.965$)
& $+0.000038$ ($p{=}0.716$)
& $-0.0189$ ($p{=}0.533$)\\
SD latent, \texttt{ramp\_early} $\gamma{=}2$ (main)
& $140$ & $\boldsymbol{-0.00042}$ ($\boldsymbol{p{=}0.271}$)
& $-0.000011$ ($p{=}0.610$)
& $\boldsymbol{+0.000008}$ ($p{=}0.669$)
& $+0.0004$ ($p{=}0.371$)\\
SDXL latent, \texttt{ramp\_early} $\gamma{=}2$ (main)
& $140$ & $+0.00006$ ($p{=}0.744$)
& $-0.000011$ ($p{=}0.652$)
& $-0.000055$ ($p{=}0.271$)
& $-0.0002$ ($p{=}0.832$)\\
\bottomrule
\end{tabular}}
\end{table}

\paragraph{Distinction from standard Gaussian rate-distortion.}
For the sampled Gaussian \NARC{} anchor alone, the direction is
consistent with an isotropic rate-distortion baseline: retaining
coordinate support under coarse scalar quantization is preferable to
discarding dimensions. The nontrivial observation is that the same
ordering persists across five qualitatively different stored-noise
cache objects. We tested DCT low-band and random-projection summaries
whose payloads are smaller than or comparable to the int8 anchor; on
SD~1.5 they lose both PSNR and LPIPS, while on SDXL they can raise
PSNR through smoothing but still substantially worsen LPIPS.
The two compression families are therefore \emph{duals}: subspace
summaries preserve fewer coordinates with higher per-coordinate
precision, whereas the noise-anchor regime favors full-dimensional
retention with reduced per-coordinate precision. The experiments show
where this direction is strong, transitional, or boundary-level across
cache objects.

\paragraph{Two mechanism conditions.}
The asymmetry reduces to two conditions on the cached object:
\begin{itemize}\setlength\itemsep{0.2em}
\item \emph{Per-dimension magnitude exceeds the quantization step.}
Otherwise even int8 incurs measurable loss --- the
\emph{boundary} regime, instantiated by PnP DirectInv's
small-magnitude implicit residuals.
\item \emph{Information is isotropically dispersed across ambient space.}
Otherwise some subspace projection catches scene structure
and partially survives --- the \emph{transition} regime, instantiated
by DDIM-Inv's deterministic $x_T^{\text{inv}}$, where reverse
deterministic dynamics imprint scene structure on the cache.
\end{itemize}
When both conditions hold we obtain \emph{strong compression asymmetry}:
full-dimensional element-wise quantization is near-lossless, whereas
support-reduced summaries fall off the PSNR/LPIPS Pareto frontier.

\paragraph{The asymmetry generalises across the stored-noise paradigm.}
We applied the same compression ladder (method-native full payload,
explicit fp16 where meaningful, int8, int4, DCT-low, random projection) to five stored-noise
inversion methods.
The direction (element-wise int8
$>$ subspace projection) holds in all five, with the magnitude
gradient predicted by the two conditions (sub-tables in
Appendix~\ref{app:f1}). NARC's $\epsstar$ (sampled Gaussian),
Edit-friendly's $T{+}1$ stored noise maps, and
LEDITS++'s~\citep{brack2024leditspp} $T$ stored noise maps land in
\emph{strong asymmetry}; DDIM-Inv's deterministic $x_T^{\text{inv}}$ lands
in \emph{transition}; PnP DirectInv's small-magnitude residuals
land in \emph{boundary}. The cache-accounted SDXL~$1024^2$ main
schedule sweep preserves the same full-support preference, but with a
metric split rather than a PSNR collapse: int8 and packed int4 remain
close to full, while support-reduced summaries have much worse LPIPS
despite sometimes higher PSNR (Appendix~\ref{app:sdxl}). The asymmetry
is therefore a paradigm-wide property of the noise latent space, not a
NARC-specific artifact.

\subsection{Finding 2: Trajectory binding and score-prior coupling}
\label{sec:mech:f2}
\label{sec:exp:f2}

The convex blend (Eq.~\ref{eq:blend}) reduces to an algebraic
identity at $\lambda_t{\equiv}1$: substituting Eq.~\ref{eq:blend} into
the DDIM step under the forward-trajectory assumption gives
$\hat x_{0,t} = x_0 + (1{-}\lambda_t)\sigma_t(\epsstar-\hateps_t)$
with $\sigma_t=\sqrt{(1{-}\bar\alpha_t)/\bar\alpha_t}$, so
$\lambda_t{\equiv}1$ exactly cancels the model term and recovers
$x_0$ at the SD VAE ceiling. This raises a natural concern: at our
operating points (the conservative reference $\lambda{=}0.7$ and the
main \texttt{ramp\_early} schedule with mean
$\bar\lambda{\approx}0.885$), is the procedure essentially a numerical
relative of this algebraic identity? Three independent measurements jointly refute this hypothesis.

\paragraph{(a) Anchor specificity.}
We use anchor specificity to mean specificity of the matched
$(\xtstar,\epsstar)$ trajectory pair, not that $\epsstar$ alone
encodes the source image. On the main \texttt{ramp\_early} schedule
(SD~1.5, cfg=7.5), the trajectory-matched forward anchor reconstructs
at $24.72$\,dB while random Gaussian / mismatched-image / shuffled /
sign-flipped correction anchors all collapse to $6.49$--$6.55$\,dB
--- a paired gap of $+18.17$ to $+18.23$\,dB, statistically
significant by paired Wilcoxon for all four
(Appendix~\ref{app:f2}). Appendix Fig.~\ref{fig:anchor_variants}
shows the same gap visually. The same pattern holds
on the conservative reference ($\lambda{=}0.7$, $+10.98$ to
$+11.08$\,dB), CIFAR ($+9.0$\,dB at $\lambda{=}0.5$), and SDXL
($+20.74$ to $+20.83$\,dB at \texttt{ramp\_early} $\gamma{=}2$).
This trajectory-pair specificity alone does not falsify the
algebraic-inversion hypothesis
(wrong-anchor collapse is also predicted algebraically for
$\lambda{<}1$); we therefore complement it with two further tests.

\paragraph{(b) Trained-score dependence at tested operating points.}
We replaced every UNet parameter with $\mathcal{N}(0,0.02^2)$ samples
(VAE/CLIP retained) and ran the same NARC sampler on the main
schedule. Reconstruction collapses from $24.72$ to $12.97$\,dB --- a
paired drop of $-11.75$\,dB on $140/140$ images, with SSIM
$0.737{\to}0.222$ and LPIPS $0.147{\to}0.757$ moving in the same
direction (all $p{=}5\!\times\!10^{-25}$, Appendix~\ref{app:f2}).
The collapse is monotone in $\lambda$ across
$\lambda\!\in\!\{0.5,0.6,0.7,0.8,0.9\}$ on the fixed-$\lambda$
comparator (paired drops $-7.90$ to $-13.39$\,dB). An algebraic
identity would be insensitive to UNet weights; these results reject a
pure algebraic-identity explanation at our tested operating points.
Under the SD~1.5 sampler, schedules, and CFG settings, replacing the
trained UNet with random weights collapses reconstruction. We do not
claim a theorem over arbitrary denoisers.

\paragraph{(c) cfg sensitivity is unimodal in $\lambda$.}
The fixed-$\lambda$ cfg sweep is unimodal: the cfg-induced PSNR drop
peaks near $\lambda\!\in\![0.6,0.7]$ at $2.60$\,dB and falls to
$0.81$\,dB at $\lambda{=}0.95$. The main \texttt{ramp\_early}
schedule (mean $\bar\lambda{\approx}0.885$) retains a positive paired
drop of $+0.694$\,dB ($p{=}1.1\!\times\!10^{-24}$, two-sided),
confirming a model-coupled regime rather than the algebraic-endpoint
band; the full sweep and paired tests are in Appendix~\ref{app:f2}.

\paragraph{Summary.}
Findings 1 and 2 jointly characterise the noise anchor as an
information-theoretic and dynamical object: the measured operating
point stores it element-wise, and at the operating points used in this
paper it cooperates with --- not replaces --- the trained score prior.
The next section translates this characterisation into a method.

\section{NARC: A Method Built on the Findings}
\label{sec:method}

\subsection{Forward noise anchoring and convex blend}
\label{sec:method:anchor}

Given $x_0$, sample $\epsstar\!\sim\!\mathcal{N}(0,I)$ once with a
deterministic per-image seed and form $\xtstar$ via
Eq.~\ref{eq:forward}. The reverse process uses the standard DDIM
($\eta{=}0$) update with classifier-free guidance
\citep{rombach2022sd}, where the per-step $\eps$-prediction
$\hateps_t^{\text{cfg}}=\hateps_t^{u}+w(\hateps_t^{c}-\hateps_t^{u})$
is replaced with $\eps_t^{\text{corr}}$ via Eq.~\ref{eq:blend}. The
main evaluation setup uses a deterministic, time-dependent schedule
$\lambda_t=\lambda(\bar\alpha_t)$ defined in
\S\ref{sec:method:schedule}. Drift-aware selective gating variants
(hard threshold and soft sigmoid) are evaluated as ablations in
Appendix~\ref{app:selective} and do not improve on the deterministic
schedule.

\paragraph{NARC as a parameterized family.}
Substituting Eq.~\ref{eq:blend} into the DDIM step under the
forward-trajectory assumption gives
$\hat x_{0,t} = x_0 + (1{-}\lambda_t)\,\sigma_t\,(\epsstar-\hateps_t)$
with $\sigma_t=\sqrt{(1{-}\bar\alpha_t)/\bar\alpha_t}$. \NARC{} is a
convex interpolation between algebraic identity
($\lambda_t{\equiv}1$) and plain DDIM ($\lambda_t{\equiv}0$): an
11-point fixed-$\lambda$ sweep
($\lambda\in\{0,0.1,\dots,1.0\}$) at SD cfg=1 confirms strict
monotone PSNR $12.75\!\to\!27.00$ with adjacent paired
$p<10^{-24}$ (Appendix~\ref{app:lambda}). The conservative
plateau-upper-bound fixed point $\lambda{=}0.7$ is retained as the
\emph{conservative reference} that anchors the Finding~2 mechanism analysis
(\S\ref{sec:mech:f2}); the schedule $\lambda(\bar\alpha_t)$ defined
below is the \emph{main reconstruction operating point}.

\subsection{Time-dependent correction schedule}
\label{sec:method:schedule}

Constant $\lambda_t{\equiv}\lambda$ ignores three properties of
the diffusion geometry. (i)~The DDIM update places ${\eps_t^{\text{corr}}}$
through the noise coefficient $\sqrt{1{-}\bar\alpha_t}$, so the
\emph{leverage} of any $\eps$-error on $x_{t-1}$ is largest at large
$t$ (early reverse, $\bar\alpha_t{\approx}0$) and vanishes at small
$t$. (ii)~The $\eps$-prediction model is least reliable at large
$t$, where $x_t$ is dominated by noise and contains little signal
about $x_0$ (\S\ref{sec:mech:f2}(b): replacing the trained UNet with
random weights collapses reconstruction by $-11.75$\,dB at the
schedule below). (iii)~The forward anchor $\epsstar$ contributes
proportionally to $\sqrt{1{-}\bar\alpha_t}$ to $x_t$: at late reverse
it adds little information about $x_0$ but actively suppresses the
model's accumulated knowledge, producing the perceptual blur seen at
fixed $\lambda{=}0.7$ (LPIPS $0.504$ on the conservative reference
setup; cf.\ Tab.~\ref{tab:main} where the schedule attains LPIPS
$0.147$).

These three observations jointly imply $\lambda_t$ should be
\emph{high at high noise} (anchor binds the trajectory when the model
is weakest and leverage is largest) and \emph{relax at low noise}
(model contributes texture once $x_0$ is partially exposed). The
simplest expression keying on the noise coefficient itself is
\begin{equation}
\lambda(\bar\alpha_t) \;=\; \lambda_{\max}
\;-\; (\lambda_{\max}{-}\lambda_{\min})\;\bar\alpha_t^{\,\gamma},
\qquad
\lambda_{\min}{=}0.70,\;\lambda_{\max}{=}0.95,\;\gamma{=}2.
\label{eq:ramp}
\end{equation}
We refer to this schedule as \texttt{ramp\_early}. Indexing on
$\bar\alpha_t$ (rather than reverse-step index) keeps $\lambda_t$
near $\lambda_{\max}$ across most of the reverse loop where
$\bar\alpha_t$ is small, transitioning toward $\lambda_{\min}$ only
in the last steps. The exponent $\gamma$ is the schedule-shape
parameter: increasing $\gamma$ keeps $\bar\alpha_t^\gamma$ smaller
through the high-noise part of the reverse trajectory, so the anchor
stays near $\lambda_{\max}$ longer and releases more sharply only
when the denoised signal is already exposed. With SD~1.5's
scaled-linear schedule and $T{=}50$
the main configuration gives mean $\bar\lambda{\approx}0.885$,
between fixed $\lambda{=}0.85$ and $\lambda{=}0.9$ in the
high-anchor but still model-coupled band validated by
\S\ref{sec:mech:f2}. A matched-mean fixed $\lambda{=}0.8852$ control
reaches only $22.99$\,dB, confirming that the placement of anchor
weight across noise levels, not only the mean $\lambda$, is essential
to the $\gamma{=}2$ operating point
(Appendix~\ref{app:matched-control}). With $\lambda_{\min}$ and
$\lambda_{\max}$ fixed by the mechanism analysis, $\gamma$ acts as
the schedule-shape parameter; the conservative $\gamma{=}1$ (linear
in $\bar\alpha_t$, mean $\bar\lambda{\approx}0.854$) is reported as
an ablation in Appendix~\ref{app:lambda}. All fixed-$\lambda{=}0.7$
rows below are conservative references or ablations, not the
proposed operating point.

\paragraph{Storage assumption (inversion-for-editing).}
Our byte accounting is for a materialized per-image inversion cache.
In the canonical real-image editing setting the source image is
retained, so the clean diffusion state $x_0$ is available directly
for pixel DDPMs or recomputed by VAE encoding for latent diffusion
models. Thus $\xtstar$ is recomputed from $x_0$ and the
stored/dequantized $\epsstar$ by Eq.~\ref{eq:forward}. The
additional cache payload is therefore the quantized $\epsstar$, not
model weights, prompt embeddings, the image file, the clean
diffusion state, the full trajectory, or per-step residuals.
Deterministic seeds are used for reproducibility and pairing;
seed-only procedural regeneration is excluded as a comparable
materialized-cache baseline. For SD~1.5 at $512^2$,
$\epsstar\in\mathbb{R}^{4\times64\times64}$ and int8 storage is
$16$\,KiB. In general the payload scales as
$C_{\mathrm{latent}}\cdot(H/s)\cdot(W/s)\cdot b/8$ bytes plus small
scale/metadata overhead. In an archival scenario where the source
image or clean diffusion state is discarded, $\xtstar$ would also
need to be stored; our storage comparisons therefore apply to the
editing scenario.

\paragraph{Justification for the proposed primitive.}
Finding~1 demonstrates that $\epsstar$ tolerates aggressive element-wise
compression, which is what makes the 16\,KiB operating point possible.
Finding~2 indicates that the correction is genuinely model-coupled at the
operating points used in this paper, which is what makes the procedure
non-trivial: \NARC{} is not algebraic inversion in disguise, and not
a stronger-anchor-is-better technique either (the trained-UNet and
cfg-sensitivity tests reject a pure algebraic-identity explanation at
the tested operating points). The two findings constrain both
\emph{what to store} (element-wise int8) and \emph{how to use it} (an
intermediate, noise-aware $\lambda_t$ with the model still in the
loop). The \texttt{ramp\_early} schedule of
\S\ref{sec:method:schedule} is the simplest implementation of this
constraint.

\subsection{Sampler implementation and cost}
\label{sec:method:alg}

Operationally, \NARC{} changes only the denoising target consumed by
the deterministic DDIM step. For each image, we sample $\epsstar$,
quantize and decode it to $\tilde{\eps}^{\star}$, rebuild
$\tilde{x}_T^{\star}$ from $x_0$ and $\tilde{\eps}^{\star}$ by
Eq.~\ref{eq:forward}, and start the reverse loop from this rebuilt
latent. At each timestep, the
sampler first computes the standard classifier-free-guided prediction
$\hateps_t^{\mathrm{cfg}}$, then evaluates the scheduled weight
$\lambda_t$ from Eq.~\ref{eq:ramp}, and finally passes
$(1{-}\lambda_t)\hateps_t^{\mathrm{cfg}}+\lambda_t\tilde{\eps}^{\star}$ to the
usual DDIM update. Thus $\gamma$ changes the time profile of the
anchor correction, not the model, optimizer, or number of denoising
evaluations. With the CFG split, \NARC{} uses $2T$ UNet calls for
$T$ reverse steps and stores only the 16\,KiB int8 anchor.

\section{Empirical Evaluation}
\label{sec:experiments}

\subsection{Setup}
\label{sec:exp:setup}

Main experiments use Stable Diffusion~1.5 \citep{rombach2022sd} on
the PIE-Bench++ \citep{ju2023pnp} \texttt{0\_random\_140} cohort ($N{=}140$).
Following the PIE-Bench++ benchmark protocol~\citep{ju2023pnp}
and subsequent real-image inversion/editing evaluations
\citep{mokady2022nulltext,garibi2024renoise,li2025dci,
wallace2022edict,bao2025freeinv,staniszewski2026taba},
we report PSNR, SSIM~\citep{wang2004ssim},
LPIPS~\citep{zhang2018lpips}, and MSE as primary conditional
round-trip-fidelity metrics.
In reconstruction tables these metrics are computed over the whole image;
in editing tables we use the corresponding benchmark background-region
variants. FID~\citep{heusel2017fid} and CLIP-I~\citep{radford2021clip}
are reported only as auxiliary cross-checks.
For mechanism cross-checks we additionally use the
CIFAR-10 $32{\times}32$ pixel-space DDPM/DDIM
\citep{ho2020ddpm,song2020ddim} at $N{=}2000$, $T{=}100$. Statistical
tests are paired Wilcoxon signed-rank by \texttt{image\_id},
two-sided for equality tests and for the main head-to-head tables;
mechanism checks use a pre-specified directional alternative only
when the relevant table or paragraph states it explicitly. We use two
operating points:
$\lambda{=}0.5$, cfg=1 for mechanism-isolation experiments
(compression Pareto, $\lambda$ monotonicity, validation checks), and
\texttt{ramp\_early} ($\lambda_{\min}{=}0.7$, $\lambda_{\max}{=}0.95$,
$\gamma{=}2$), cfg=7.5, after-CFG for the main reconstruction
comparisons (Eq.~\ref{eq:ramp}, \S\ref{sec:method:schedule}); for the
trajectory-binding analyses of \S\ref{sec:mech:f2} we additionally
retain the fixed $\lambda{=}0.7$ as a conservative reference
operating point to keep the cfg plateau argument
(Tab.~\ref{tab:p5}) intact. Both setups appear in the int8 vs.\ fp16
verification (Tab.~\ref{tab:int8-lossless}), so the compression
behavior is shown to be schedule-invariant. NFE is reported as the
total number of UNet forward calls; detailed per-method accounting is
given in Appendix~\ref{app:hp}. Storage assumes the
inversion-for-editing scenario (\S\ref{sec:method}). 
FID~\citep{heusel2017fid} uses Inception-v3 ($d{=}2048$) with
PIE-Bench~140 originals as reference; CLIP-I uses CLIP image
embeddings~\citep{radford2021clip}.

\subsection{Cache-accounted reconstruction trade-off}
\label{sec:exp:main}

The two findings of \S\ref{sec:mechanism} predict a specific
operating point: a 16\,KiB anchor stored element-wise (Finding~1)
combined with a model-coupled noise-aware schedule
(\texttt{ramp\_early} $\gamma{=}2$, motivated by Finding~2).
Tab.~\ref{tab:main} measures whether this operating point is
competitive against existing inversion methods on PIE-Bench++.
We measure whole-image reconstruction PSNR for all methods to isolate
inversion fidelity from any editing engine.

\begin{table}[t]
\centering
\caption{\textbf{Main reconstruction comparison under cache-accounted
stored-noise inversion.} PIE-Bench++ SD~1.5, cfg=7.5, $N{=}140$,
paired by image\_id; metrics are whole-image PSNR/SSIM/LPIPS/MSE.
$\Delta$PSNR is NARC int8 minus each baseline (fp16 minus int8 for
the fp16 row). Storage denotes additional materialized inversion cache
in the retained-source editing setting of \S\ref{sec:method}, not
standalone archival compression. The $\lambda_t{\equiv}1$ endpoint is
only a reconstruction ceiling reference because it cancels the model
term, not the proposed model-coupled operating point.}
\label{tab:main}
\small
\setlength{\tabcolsep}{3pt}
\renewcommand{\arraystretch}{1.05}
\resizebox{\textwidth}{!}{%
\begin{tabular}{lrlrrrrl}
\toprule
Method & NFE & Storage & PSNR\,$\uparrow$ & SSIM\,$\uparrow$ & LPIPS\,$\downarrow$ & MSE\,$\downarrow$ ($\!\times\!10^{-3}$) & $\Delta$PSNR \\
\midrule
\multicolumn{8}{l}{\emph{Group 1 --- Proposed (\texttt{ramp\_early} $\gamma{=}2$, $\lambda_t\in[0.70,0.95]$, mean $\bar\lambda{\approx}0.885$):}}\\
\textbf{NARC int8 (proposed)} & \textbf{100} & \textbf{$\epsstar$ int8, 16\,KiB} & \textbf{24.724} & \textbf{0.737} & \textbf{0.147} & \textbf{4.46} & --- \\
NARC fp16 (anchor full)       & 100 & 32\,KiB           & 24.724 & 0.737 & 0.147 & 4.45 & $+0.0004$ (n.s.)\\
\midrule
\multicolumn{8}{l}{\emph{Group 2 --- Modern non-exact baselines (different mechanisms; paired head-to-head):}}\\
DDIM Inversion\,\citep{song2020ddim}             & 150 & $x_0$ access     & 14.93 & 0.542 & 0.498 & 36.5 & $\boldsymbol{+9.79}$ ($p{=}10^{-24}$)\\
ReNoise $K{=}5$\,\citep{garibi2024renoise}       & 400 & iterative        & 15.09 & 0.550 & 0.466 & 34.8 & $\boldsymbol{+9.63}$ ($p{=}10^{-24}$)\\
DCI $K{=}5$ (default)\,\citep{li2025dci}         & 212 & $z_0$, 64\,KB    & 15.33 & 0.555 & 0.451 & 32.9 & $\boldsymbol{+9.40}$ ($p{=}10^{-24}$)\\
FreeInv \citep{bao2025freeinv}     & 150 & $x_T{+}$angles, 32\,KB & 16.96 & 0.627 & 0.332 & 23.0 & $\boldsymbol{+7.77}$ ($p{=}10^{-24}$)\\
taba \citep{staniszewski2026taba} ($\text{fbt}{=}2$) & 148 & $x_T{+}$seed, 32\,KB & 15.06 & 0.546 & 0.468 & 35.3 & $\boldsymbol{+9.66}$ ($p{=}10^{-24}$)\\
\midrule
\multicolumn{8}{l}{\emph{Group 3 --- Stored-noise methods within the same paradigm (NARC = extreme compression point):}}\\
PnP DirectInv\,\citep{ju2023pnp} & 250 & ddim$+$offset, 6.4\,MB & 21.48 & 0.689 & 0.212 & 11.1 & $\boldsymbol{+3.24}$ ($p{=}7.4{\times}10^{-16}$)\\
Edit-friendly DDPM\,\citep{huberman2024editfriendly}        & 200 & $T{+}1$ maps, 3.2\,MB  & 26.70 & 0.777 & 0.066 & 3.34 & $-1.97$ ($p{=}2.3{\times}10^{-23}$)\\
\midrule
\multicolumn{8}{l}{\emph{Group 4 --- Other paradigms (PSNR ceiling references):}}\\
Null-text\,\citep{mokady2022nulltext} $K{=}10$ ($N{=}50$) & 609 & embeddings, 11.5\,MB & 25.98 & 0.760 & 0.100 & 4.15 & $-1.16$ ($p{=}5.6{\times}10^{-5}$)\\
EDICT\,\citep{wallace2022edict}              & 400 & dual stream, 32\,KB  & 26.98 & 0.780 & 0.063 & 3.22 & $-2.25$ ($p{=}10^{-24}$)\\
\bottomrule
\end{tabular}}
\end{table}

\paragraph{Degenerate ceiling reference.}
The $\lambda_t{\equiv}1$ endpoint is not included as a method row
because it cancels the model term and serves only as a reconstruction
ceiling; Appendix~\ref{app:lambda} reports the SD VAE ceiling at
$27.00$\,dB. The proposed row is instead the model-coupled
\texttt{ramp\_early} operating point evaluated against existing
inversion methods under the cache-accounted protocol above.

\paragraph{Modern non-exact baselines.}
\NARC{} (\texttt{ramp\_early} $\gamma{=}2$, int8 anchor) is paired
$+7.77$ to $+9.79$\,dB above DDIM-Inv, ReNoise, DCI, FreeInv, and
taba on PSNR (all paired Wilcoxon $p\!\leq\!10^{-24}$, the test
floor at $N{=}140$), with proportionally large LPIPS gains
($\Delta{=}{-}0.185$ to $-0.351$, all $p\!\leq\!2{\times}10^{-24}$).
\NARC{} simultaneously uses $32$--$75\%$ fewer UNet calls. Among
cache-bearing baselines, its 16\,KiB payload is smaller, with the
largest storage advantage appearing against stored-noise
methods within the same paradigm, such as PnP DirectInv. The PSNR ranking is mirrored on MSE
(NARC $4.46{\times}10^{-3}$ vs.\ $23.0$--$36.5{\times}10^{-3}$) and
SSIM (NARC $0.737$ vs.\ $0.542$--$0.627$).

\paragraph{Stored-noise methods within the same paradigm.}
\texttt{ramp\_early} $\gamma{=}2$ \emph{reverses the performance gap relative to PnP DirectInv}: paired $+3.24$\,dB ($p{=}7.4{\times}10^{-16}$) and
$\Delta$LPIPS $=-0.065$ ($p{=}6.7{\times}10^{-8}$) at about
$400\times$ less storage and $60\%$ fewer UNet calls. Edit-friendly DDPM
(3.2\,MB, $T{+}1$ stored noise maps) retains the paradigm PSNR
ceiling at $-1.97$\,dB above \NARC{}, but the main schedule reduces
this gap by more than $4\times$ relative to the conservative fixed
$\lambda{=}0.7$
configuration ($-9.10$\,dB). Finding~1 predicts that the same int8
compression is not the dominant source of the residual Edit-friendly
gap (Appendix~\ref{app:f1}); the remaining difference is primarily
one of paradigm operating point rather than representational
efficiency.

\paragraph{Other paradigms.}
EDICT (algebraic exact-invertibility via dual coupled streams) and
Null-text (per-image optimization) remain $1.3$--$2.5$\,dB above
\NARC{}, but require roughly $4$--$6\times$ more UNet calls and
$2$--$700\times$ larger materialized caches, depending on the
paradigm. Relative to the fixed-$\lambda{=}0.7$ reference, where the
gaps were roughly $-8.4$\,dB to Null-text and $-9.4$\,dB to EDICT,
the main schedule reduces them to $-1.16$ and $-2.25$\,dB. We do not
claim to surpass these ceilings,
which sit in different paradigms with different cost structures.

\subsection{Editing application demo}
\label{sec:exp:editing}

We apply \NARC{} to the PIE-Bench++ mask-rich subset
($N{=}140$, cfg=7.5) and compare \texttt{no\_narc}, whole-image
blending, mask-background blending, and mask-foreground blending.
Whole-image blending gives the best raw background fidelity but leaks
into the foreground and degrades target alignment. The main
\texttt{mask\_bg} variant keeps target alignment close to the
no-NARC baseline ($\Delta$CLIP-T $=-0.009$ on SD~1.5, $-0.010$ on
SDXL) while improving PSNR by about $+5.1$\,dB and reducing
background LPIPS by $0.29$--$0.31$ on both backbones; full
per-variant tables and the \texttt{mask\_fg} negative control are in
Appendix~\ref{app:edit}.

\section{Discussion and Limitations}
\label{sec:discussion}

\paragraph{Choice of metrics.}
We follow the inversion-fidelity convention of PIE-Bench++, DCI,
EDICT, ReNoise, SwiftEdit, FreeInv, and taba and report
PSNR/SSIM/LPIPS/MSE --- the four PIE-Bench background-preservation
metrics. FID measures \emph{marginal}
distribution similarity and CLIP-I measures \emph{global semantic
identity}, whereas inversion fidelity is the \emph{conditional,
per-image round-trip}. For transparency, FID (Inception-v3,
$N{=}140$) and CLIP-I cross-checks are in
Appendices~\ref{app:fid} and~\ref{app:clip}; both are best read as
auxiliary diagnostics rather than head-to-head reconstruction
metrics.

\paragraph{Operating point vs.\ algebraic identity endpoint.}
\NARC{} contains an algebraic identity as a \emph{degenerate
endpoint}: when $\lambda_t{\equiv}1$, Eq.~\ref{eq:blend} reduces to
$\eps_t^{\text{corr}}{=}\epsstar$, and \emph{under the
forward-trajectory assumption} the reverse update recovers the SD
VAE reconstruction ceiling at $27.00$\,dB
(Appendix~\ref{app:lambda}). The \texttt{ramp\_early} schedule
(Eq.~\ref{eq:ramp}) is deliberately away from this limit: its mean
$\bar\lambda{\approx}0.885$ at $\gamma{=}2$ remains below endpoint
comparators, and its small-$t$ endpoint is $\lambda{=}0.70$ rather
than $1$. Three measurements show this is a non-degenerate,
model-coupled regime: randomizing the UNet collapses reconstruction
by $-11.75$\,dB ($p{=}5\!\times\!10^{-25}$); the cfg sweep retains a
paired $+0.694$\,dB model-coupled drop; and a mean-$\lambda$ matched
fixed control reaches only $22.99$\,dB vs.\ $24.72$\,dB for
\texttt{ramp\_early} $\gamma{=}2$ (Appendix~\ref{app:matched-control}).
Together, these tests place \NARC{} in a model-coupled
anchor-correction regime rather than an exact-inversion endpoint.

\paragraph{Paradigm framing.}
We organize methods into four paradigms --- \emph{stored-noise}
(\NARC{}, Edit-friendly, PnP DirectInv), \emph{exact-inversion}
(EDICT), \emph{optimization-based} (Null-text), and \emph{modern
non-exact} (DDIM-Inv, ReNoise, DCI, FreeInv, taba). \NARC{} is the
extreme-compression operating point of the stored-noise paradigm. It
outperforms the evaluated modern non-exact baselines and PnP DirectInv
under our measured protocol, but we do not claim to surpass
large-cache ceilings such as Edit-friendly DDPM or the
exact/optimization paradigms represented by EDICT and Null-text.

\paragraph{Limitations.}
(i) int8 is the only element-wise scheme we treat as the paper's
validated operating point. Packed int4 survives the cache-accounted
SD/SDXL F1 audits, but we have not revalidated it across the full
main reconstruction/editing suite, so it remains an ablation rather
than the proposed point. (ii) Support-reduced summaries should not be
summarized as universal PSNR collapse: on SDXL and SD~3 they can raise
PSNR/SSIM through smoothing, while still worsening LPIPS and falling
off the perceptual Pareto frontier. (iii) Full empirical transfer to
DiT-based backbones remains future work; Appendix~\ref{app:sd3}
reports a suggestive SD~3/MMDiT probe rather than a main DiT-transfer
claim. (iv) The 16\,KiB storage
claim is additional-cache accounting for retained-source editing; if
the source image or clean diffusion state is discarded, $\xtstar$ or
equivalent information must also be stored. (v) A five-seed audit
shows small anchor-seed variability (Appendix~\ref{app:seed}); deeper
integration with editing frameworks such as P2P/MasaCtrl remains
future work.

\section{Conclusion}
\label{sec:conclusion}

We isolated two mechanism findings governing real-image diffusion
inversion via cached forward noise. First, diffusion noise anchors
exhibit an \emph{element-wise vs.\ subspace} compression asymmetry:
retaining all coordinates at low precision is far more faithful than
low-dimensional projection across five stored-noise inversion
methods.
Second, successful reconstruction at the tested operating points is
not an algebraic identity in disguise: it relies on a
trajectory-matched forward anchor and the trained score model, placing
the method in a model-coupled anchor-correction regime.
On these foundations, \NARC{} is the minimal training-free primitive
that turns the findings into an operating point --- store one int8
forward-noise anchor and reuse it through the scheduled blend
$\lambda(\bar\alpha_t){=}\lambda_{\max}{-}(\lambda_{\max}{-}\lambda_{\min})\bar\alpha_t^{\gamma}$
with $(\lambda_{\min},\lambda_{\max},\gamma){=}(0.7,0.95,2)$.
The resulting ``high anchor at high noise, model-driven texture at
low noise'' schedule establishes a 16\,KiB extreme-compression point
within the stored-noise paradigm on PIE-Bench++ SD~1.5 at $512^2$ ---
\NARC{} is paired $+7.77$ to
$+9.79$\,dB above five modern non-exact baselines and $+3.24$\,dB
above PnP DirectInv (6.4\,MB), at $32$--$75\%$ fewer UNet calls. The
central contribution is the mechanism-to-method link: the findings
explain which cache object to store and how to couple it back into the
reverse trajectory.

\section*{Acknowledgments}
This work was partly supported by Institute of Information \& communications
Technology Planning \& Evaluation (IITP) grant funded by the Korea government
(MSIT) (RS-2026-25528270, DevOps-based Post Quantum Cryptography Switching
Automation Technology and Open Platform Technology, 50\%) and Institute of
Information \& communications Technology Planning \& Evaluation (IITP) grant
funded by the Korea government (MSIT) (RS-2026-25530181, Development of
Inter-Domain Quantum Security System Combining PQC and QKD, 50\%).


\bibliographystyle{plainnat}
\bibliography{references}

\clearpage
\appendix

\section{Qualitative galleries and compression plot}
\label{app:galleries}

\subsection{Trajectory-pair specificity gallery}

\begin{figure}[H]
\centering
\includegraphics[width=0.82\textwidth,height=0.30\textheight,keepaspectratio]{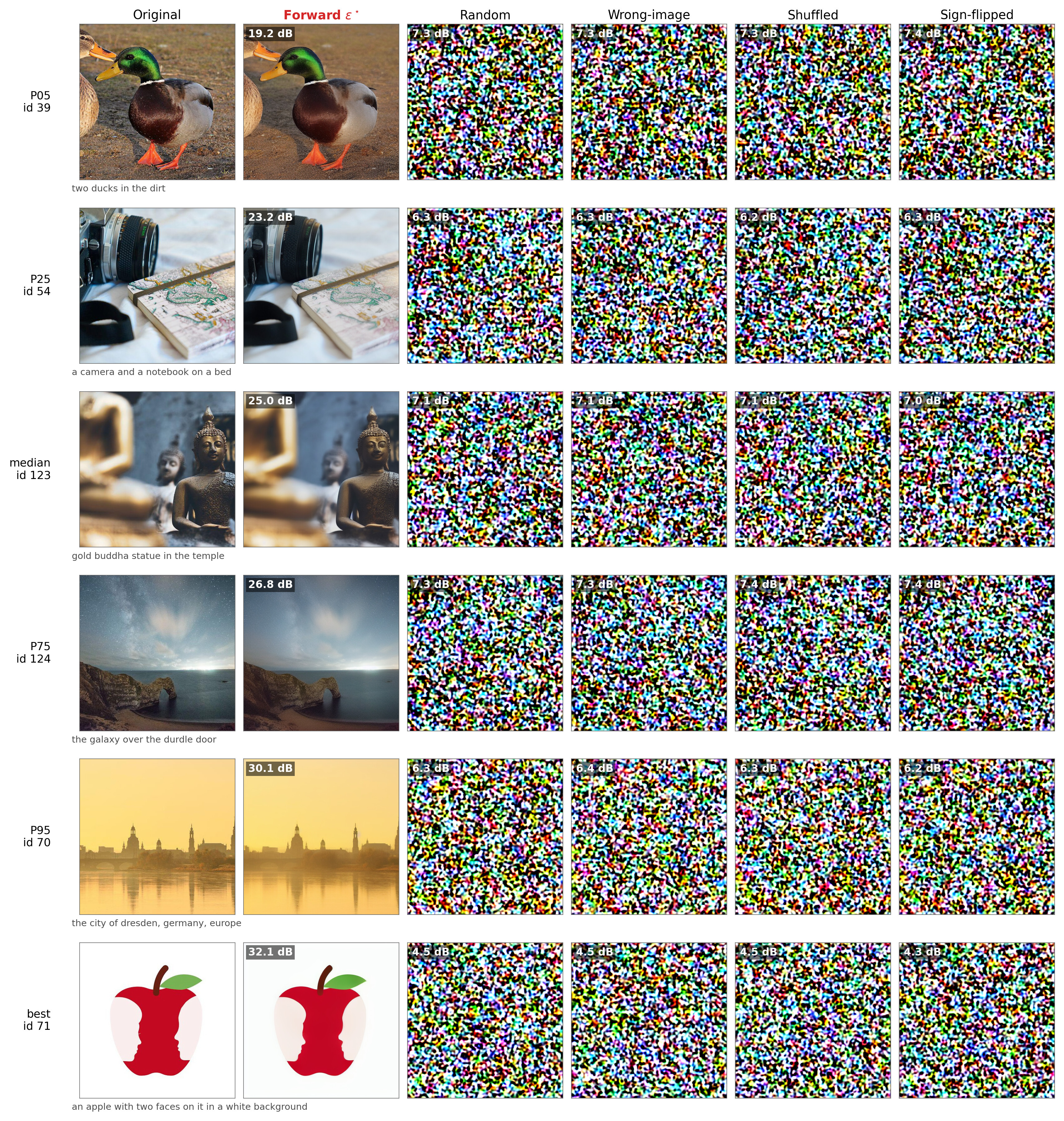}
\caption{\textbf{Visual trajectory-pair specificity.} Same images reconstructed
with the forward anchor and four perturbations under the main
\texttt{ramp\_early} $\gamma{=}2$ schedule. Only the
trajectory-matched forward anchor preserves the source trajectory,
matching the paired
$+18.17$--$18.23$\,dB gap in Appendix~\ref{app:f2}.}
\label{fig:anchor_variants}
\end{figure}

\subsection{Reconstruction gallery}

\begin{figure}[H]
\centering
\includegraphics[width=0.90\textwidth,height=0.40\textheight,keepaspectratio]{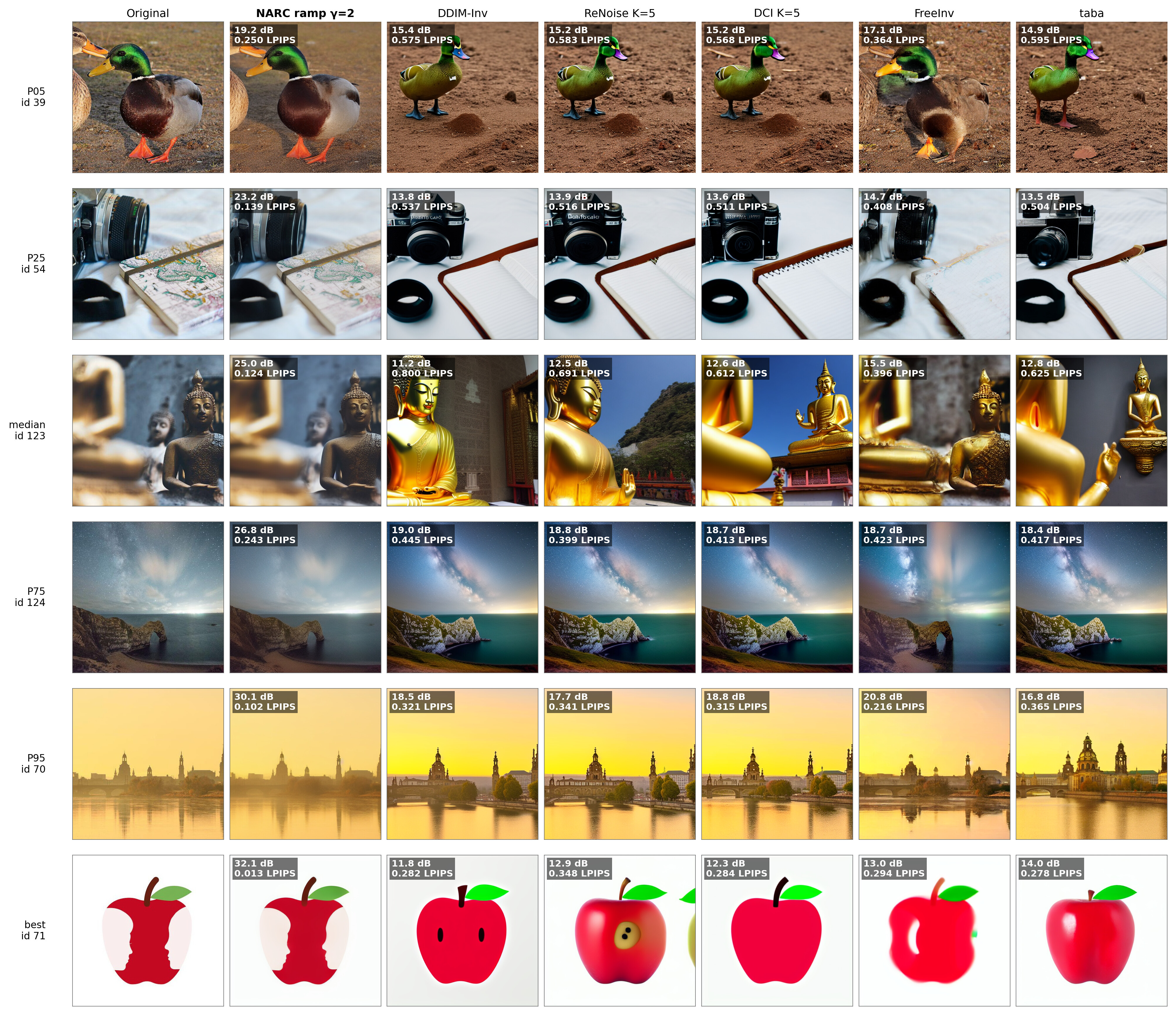}
\caption{\textbf{Reconstruction gallery.} Representative PIE-Bench++
examples comparing the original image, \NARC{} \texttt{ramp\_early}
$\gamma{=}2$, and five modern non-exact baselines. The visual gaps
match Tab.~\ref{tab:main}: \NARC{} preserves the source layout while
baselines often drift.}
\label{fig:recon_gallery}
\end{figure}

\subsection{Storage-quality plot}

\begin{figure}[H]
\centering
\includegraphics[width=0.78\textwidth]{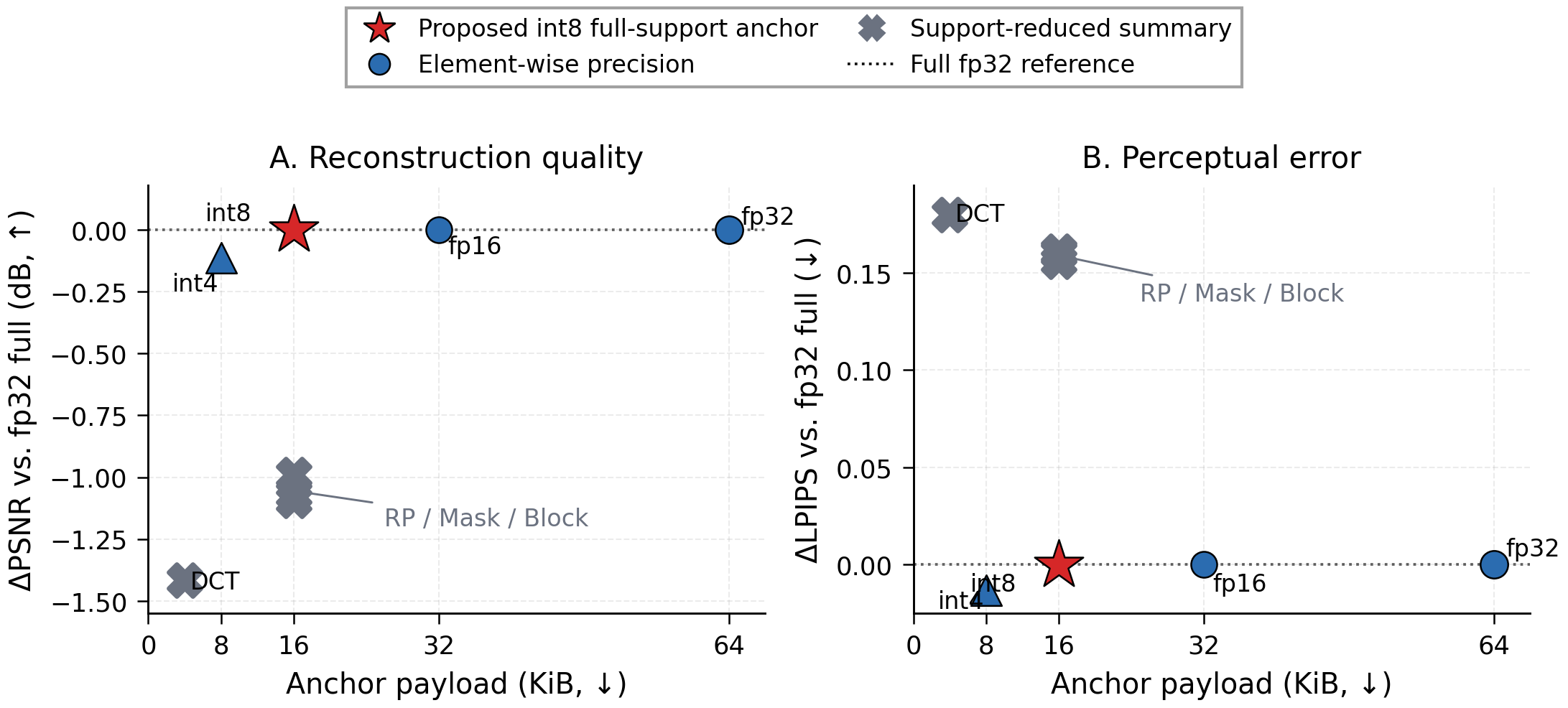}
\caption{\textbf{Anchor compression asymmetry on SD~1.5.}
Payload is shown in KiB for the explicit SD~1.5 $512^2$ fp32-runtime
audit, with the reverse start rebuilt from each decoded payload.
Axes report changes from the fp32 full-anchor reference; higher
$\Delta$PSNR and lower $\Delta$LPIPS are better. Element-wise
fp16/int8 preserve reconstruction quality, packed int4 remains close,
and support-reduced DCT, random-projection, spatial-mask, and
block-average summaries are much worse at smaller or comparable
payloads. The no-anchor reference is $9.42$\,dB / $0.815$ LPIPS and
is omitted from the axes because it is not a compression method.}
\label{fig:storage_quality}
\end{figure}

\subsection{Extended reconstruction gallery}

\begin{figure}[H]
\centering
\includegraphics[width=\textwidth]{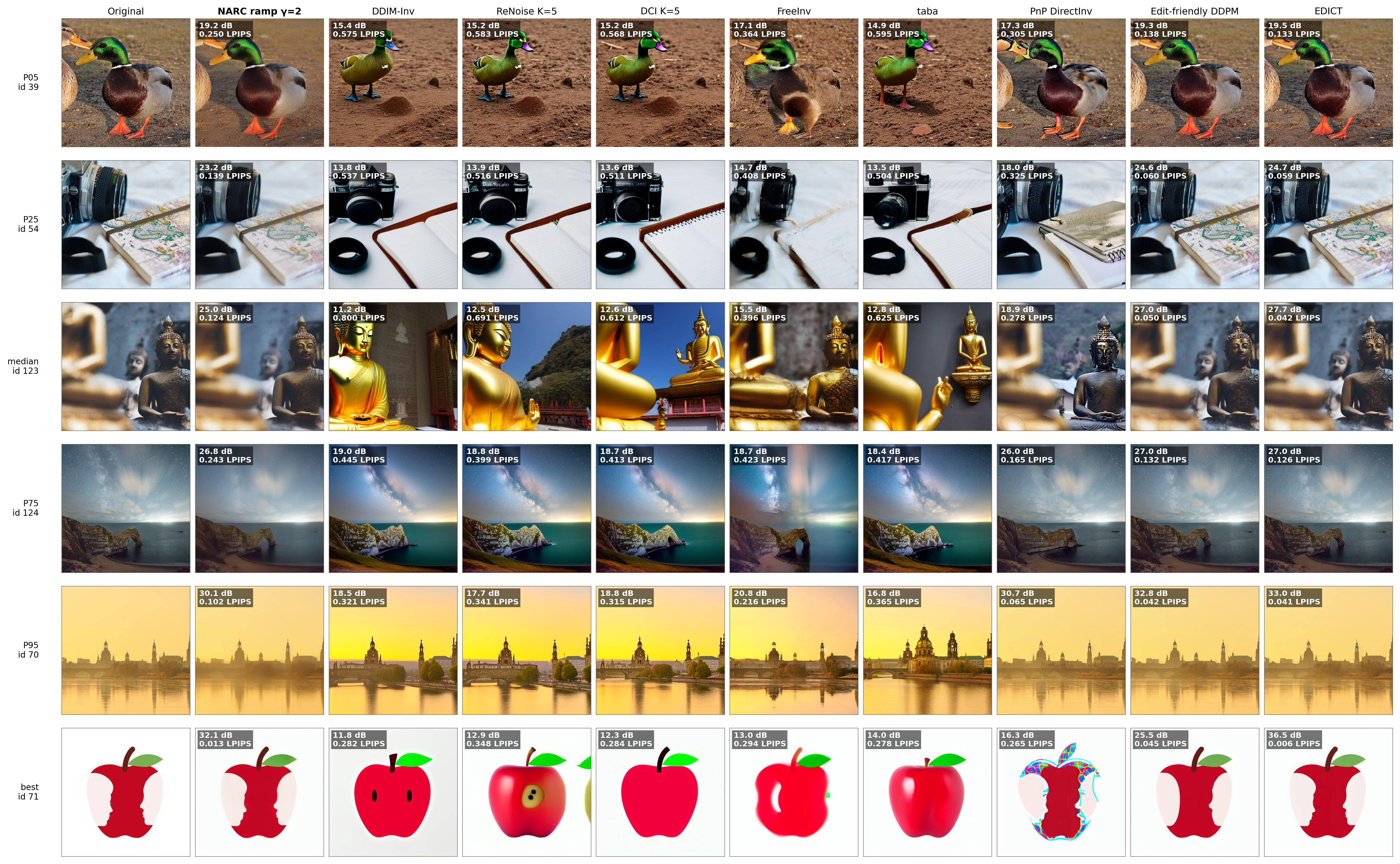}
\caption{\textbf{Extended reconstruction gallery.} Additional
PIE-Bench++ examples for the visual comparison summarized in
Fig.~\ref{fig:recon_gallery}.}
\label{fig:recon_gallery_full}
\end{figure}

\section{Distribution-realism cross-check (FID)}
\label{app:fid}

\subsection{Rationale for treating FID as a secondary metric}

We follow the inversion-fidelity convention of PIE-Bench++
\citep{ju2023pnp}, DCI~\citep{li2025dci},
EDICT~\citep{wallace2022edict},
ReNoise~\citep{garibi2024renoise},
Null-text~\citep{mokady2022nulltext},
SwiftEdit~\citep{nguyen2025swiftedit},
FreeInv~\citep{bao2025freeinv}, and
taba~\citep{staniszewski2026taba} --- none of which adopts FID as a
primary reconstruction metric. The reason is structural: real-image
inversion is a \emph{conditional} per-image round-trip task
(``how faithfully is this one source $x_0$ reconstructed?''),
while FID measures the \emph{marginal} similarity between two image
distributions; the two axes can be negatively correlated by design
(\S\ref{sec:discussion}). In addition, FID with $N{=}140$ samples is
in the small-$N$ noisy regime --- standard practice uses $\geq\!2$K
samples for stable estimates. We therefore report Inception-v3
($d{=}2048$) FID against PIE-Bench~140 originals only as a
distribution-realism cross-check; the same direction as PSNR is
preserved within each paradigm, but absolute numbers across
paradigms must be read with the trade-off below in mind.

\subsection{FID cross-check for the main reconstruction table}

The main scheduled NARC FID row in this appendix uses the
cache-accounted int8 \texttt{ramp\_early} $\gamma{=}2$ samples from
Tab.~\ref{tab:main}; the $\gamma{=}1$ row remains a schedule-ablation
cross-check. Under this accounting, NARC's FID drops from $146.88$
at the full-anchor fixed-$\lambda{=}0.7$ reference to
$\boldsymbol{43.32}$ at \texttt{ramp\_early} $\gamma{=}2$ (and
$55.89$ at the $\gamma{=}1$ ablation), a $\boldsymbol{-103.6}$\,FID
improvement that mirrors the $+7.13$\,dB PSNR / $-0.36$ LPIPS
quality jump. The schedule's FID is $64.1$ below FreeInv (the
strongest modern non-exact baseline on FID), placing it well below
the modern-non-exact FID band, while still above the
FID$\approx20$ ceiling represented by Edit-friendly DDPM and EDICT,
which pay different costs: a much larger stored-noise cache or an
exact-inversion dual-stream formulation.

\begin{table}[H]
\centering\small
\caption{\textbf{Distribution-realism cross-check.} Inception-v3
($d{=}2048$) FID on PIE-Bench++ $N{=}140$ originals as reference.
Same cohort and methods as Tab.~\ref{tab:main}. Read as
\emph{secondary}, not primary, evidence (small-$N$ caveat).}
\label{tab:fid-cross}
\begin{tabular}{lrr}
\toprule
Method & PSNR & FID ($N{=}140$, Inception-v3) \\
\midrule
\textbf{NARC \texttt{ramp\_early} $\gamma{=}2$ int8 (proposed)} & \textbf{24.72} & \textbf{43.32}\\
NARC \texttt{ramp\_early} $\gamma{=}1$ (ablation)   & 23.88 & 55.89\\
NARC int8, fixed $\lambda{=}0.7$ (reference)        & 17.59 & 146.03\\
NARC fp16, fixed $\lambda{=}0.7$ (reference)        & 17.60 & 146.88\\
\midrule
DDIM Inversion                             & 14.93 & 136.86\\
ReNoise $K{=}5$                            & 15.09 & 132.33\\
DCI $K{=}5$                                & 15.33 & 131.61\\
FreeInv                                    & 16.96 & 107.37\\
taba ($\text{fbt}{=}2$)                    & 15.06 & 134.17\\
\midrule
PnP DirectInv (same paradigm, 6.4\,MB)     & 21.48 & 86.60\\
Edit-friendly DDPM (same paradigm, 3.2\,MB) & 26.70 & \textbf{20.08}\\
\midrule
Null-text $K{=}10$ ($N{=}50$)              & 25.98 & 136.15\\
EDICT (exact-inversion paradigm)           & 26.98 & 18.92\\
\bottomrule
\end{tabular}
\end{table}

\subsection{Interpretation of the FID results}

At the conservative reference $\lambda{=}0.7$,
NARC's FID was $146$, well above FreeInv's $107$ (best modern
non-exact). In the cache-accounted int8 main run for
\texttt{ramp\_early} $\gamma{=}2$, FID drops to $43.32$, achieving
lower FID than every modern non-exact baseline on this auxiliary
axis ($-64.1$ vs.\
FreeInv, $-88.3$ vs.\ DCI K=5). Against stored-noise methods within
the same paradigm,
the residual gap to PnP DirectInv ($86.6$) flips from $-60$ to
$+43.3$\,FID in NARC's favour under the same cross-check, while
Edit-friendly DDPM and EDICT remain at the joint PSNR$/$FID ceiling,
but with different costs: Edit-friendly uses a much larger
stored-noise cache, whereas EDICT uses an exact-inversion dual-stream
formulation with higher NFE. We
therefore retain FID as a secondary cross-check; the
\emph{conditional} per-image PSNR/SSIM/LPIPS/MSE remain the primary
inversion-fidelity metrics.

\subsection{Direct int8 FID audit}

As an auxiliary distribution-realism check, direct-measurement
anchor-compression runs in conservative setups give
$\Delta$FID $=-3.27$ on CIFAR pixel $L_{0.5}$, $+0.92$ on SD latent
$L_{0.5}$, and $-0.62$ on SD latent fixed $L_{0.7}$. These numbers
are not used as headline evidence for the cache-accounted
compression claim in Tab.~\ref{tab:int8-lossless}; they simply show
that the int8/full-anchor change is small relative to the
small-$N$ FID noise in conservative reference setups.

\subsection{Relation to Finding~1}

We do not use FID as the primary evidence for Finding~1. The
cache-accounted F1 audits in
Fig.~\ref{fig:storage_quality}, Appendix~\ref{app:f1}, and
Appendix~\ref{app:sdxl} are paired conditional reconstructions, where
PSNR/SSIM/LPIPS/MSE are the field-standard axes. This matters because
the SDXL and SD~3 compression pilots show a metric split: some
support-reduced summaries raise PSNR/SSIM through smoothing while
still worsening LPIPS. We therefore keep FID as a distribution-realism
cross-check for the main reconstruction table, not as a headline
compression metric.

\section{Auxiliary semantic similarity (CLIP-I cross-check)}
\label{app:clip}

\subsection{Definition}

CLIP-I is the cosine similarity in the CLIP
image-encoder feature space between the source image $x_0$ and the
reconstructed image $\hat x_0$
($\text{CLIP-I}=\cos(f_{\text{img}}(x_0),f_{\text{img}}(\hat x_0))$;
higher is better). It captures \emph{global semantic identity}
rather than pixel- or patch-level fidelity, and is therefore
complementary to PSNR/SSIM/LPIPS/MSE.

\subsection{Rationale for treating CLIP-I as an auxiliary metric}

\emph{(i)}~The inversion-fidelity benchmarks we follow ---
PIE-Bench++ \citep{ju2023pnp}, DCI \citep{li2025dci},
ReNoise \citep{garibi2024renoise}, EDICT \citep{wallace2022edict},
SwiftEdit \citep{nguyen2025swiftedit}, FreeInv \citep{bao2025freeinv}
--- do not adopt CLIP-I as a primary reconstruction metric;
they use PSNR/SSIM/LPIPS/MSE (and CLIP-T/CLIP-direction in editing
contexts, not CLIP-I in reconstruction).
\emph{(ii)}~NARC's contribution is the \emph{conditional} round-trip
fidelity at extreme storage compression (Tab.~\ref{tab:main}), not
high-level semantic identity at arbitrary cost.
\emph{(iii)}~Paradigms with caches $200$--$700\times$ larger than
NARC's 16\,KiB (Edit-friendly, PnP, Null-text), or with algebraic
exact-invertibility (EDICT), trivially reach high CLIP-I; placing
CLIP-I in the main table conflates the storage--quality and
paradigm-difference axes already discussed in
\S\ref{sec:exp:main} and \S\ref{sec:discussion}.

\begin{table}[H]
\centering\small
\caption{\textbf{Auxiliary CLIP-I cross-check}
(PIE-Bench++ SD~1.5 cfg=7.5, $N{=}140$). Higher is better. Same
cohort and baseline methods as Tab.~\ref{tab:main}; the NARC row is
the cache-accounted \texttt{ramp\_early} $\gamma{=}2$ int8 main run.}
\label{tab:clip-cross}
\begin{tabular}{lrr}
\toprule
Method & Storage & CLIP-I\,$\uparrow$ \\
\midrule
\multicolumn{3}{l}{\emph{Group 1 --- Proposed:}}\\
NARC \texttt{ramp\_early} $\gamma{=}2$ int8 & 16\,KiB & 0.9538 \\
\midrule
\multicolumn{3}{l}{\emph{Group 2 --- Modern non-exact baselines:}}\\
DDIM Inversion                     & $x_0$ access  & 0.8379 \\
ReNoise $K{=}5$                    & iterative     & 0.8443 \\
DCI $K{=}5$                        & 64\,KB        & 0.8513 \\
taba ($\text{fbt}{=}2$)            & 32\,KB        & 0.8470 \\
FreeInv                            & 32\,KB        & 0.8911 \\
\midrule
\multicolumn{3}{l}{\emph{Group 3 --- Stored-noise methods within the same paradigm ($200$--$400\times$ larger cache):}}\\
PnP DirectInv                      & 6.4\,MB       & 0.9126 \\
Edit-friendly DDPM                 & 3.2\,MB       & 0.9679 \\
\midrule
\multicolumn{3}{l}{\emph{Group 4 --- Other paradigms:}}\\
Null-text $K{=}10$ ($N{=}50$)      & 11.5\,MB      & 0.9653 \\
EDICT (exact-inversion)            & 32\,KB        & \textbf{0.9693} \\
\bottomrule
\end{tabular}
\end{table}

\subsection{Interpretation of the CLIP-I results}

With the cache-accounted main schedule,
NARC's CLIP-I is $0.9538$: it is above the modern non-exact baselines
and PnP DirectInv, but still below the large-cache or exact/optimization
references (Edit-friendly DDPM, Null-text, EDICT). This is consistent
with the storage--quality and paradigm-difference axes already analyzed
for PSNR and FID. NARC therefore does not claim CLIP-I superiority; it
claims the \emph{16\,KiB additional-cache operating point}
on the PSNR/SSIM/LPIPS/MSE axes (Tab.~\ref{tab:main}), with CLIP-I used
only as an auxiliary diagnostic.

\section{Supplementary tables for F1 (per-method)}
\label{app:f1}

In the method-transfer tables below, \texttt{M0\_full} denotes the
uncompressed cache in that method's native runtime dtype, not always
an fp32 tensor. SD \NARC{}, Edit-friendly DDPM, and PnP DirectInv use
fp32 full caches in our runs, whereas LEDITS++ and DDIM-Inv use
native fp16-runtime caches. Storage ratios are therefore most
meaningful within each table; the SD \NARC{} table additionally
reports explicit KiB payloads.
Each subsection below names the cached object being compressed; the
corresponding table then reports the element-wise and support-reduced
compression variants for that object.

\subsection{SD~1.5 \NARC{} anchor}
\label{app:f1-sd-narc}

This is the paper's direct anchor-compression audit at the
$\lambda{=}0.5$, cfg=1 mechanism setup.
Storage ratios are measured against an fp32 materialized anchor. For
each compressed row, $\xtstar$ is rebuilt from the decoded
$\epsstar$ payload before reverse sampling, so the byte accounting
matches the actual inversion cache. The int4 row below is packed
4-bit storage.
\begin{table}[H]
\centering\small
\caption{\textbf{SD~1.5 \NARC{} $\epsstar$ compression.}
Cache-accounted fp32-runtime mechanism setup ($\lambda{=}0.5$,
cfg=1, $N{=}140$). Payloads are explicit for SD~1.5 $512^2$
latents.}
\label{tab:f1-sd-narc}
\begin{tabular}{lrrrrr}
\toprule
Scheme & Payload & Storage ratio & PSNR & LPIPS & paired $\Delta$PSNR\\
\midrule
M0\_full (fp32)             & 64\,KiB & 1.000 & 16.7224 & 0.5760 & ---\\
M1\_fp16                    & 32\,KiB & 0.500 & 16.7224 & 0.5760 & $+0.0000$ (n.s.)\\
\textbf{M2\_int8 (proposed)}& \textbf{16\,KiB} & \textbf{0.250} & \textbf{16.7225}
& \textbf{0.5759} & $\boldsymbol{+0.0001}$ (n.s.)\\
M3p\_packed int4            & 8\,KiB  & 0.125 & 16.609 & 0.563 & $-0.114$\\
M4\_DCT low (cutoff=16)     & 4\,KiB  & 0.0625 & 15.308 & 0.756 & $-1.415$\\
M5\_random projection       & 16\,KiB & 0.250 & 15.654 & 0.736 & $-1.069$\\
M7\_spatial mask            & 16\,KiB & 0.250 & 15.628 & 0.737 & $-1.094$\\
M8\_block average           & 16\,KiB & 0.250 & 15.733 & 0.732 & $-0.989$\\
M6\_no\_narc                & 0\,KiB  & 0.000 & 9.42  & 0.815 & $-7.30$\\
\bottomrule
\end{tabular}
\end{table}

\subsection{Edit-friendly DDPM noise maps}
\label{app:f1-editfriendly}

The compressed object is the method's stored $T{+}1$ noise-map cache
($N{=}140$ paired, SD~1.5 fp32, $T{=}50$, $\eta{=}1$, cfg=1).
\begin{table}[H]\centering\small
\caption{\textbf{F1 transfer to Edit-friendly DDPM noise maps.}
\texttt{M0\_full} is the uncompressed fp32 cache of $T{+}1$ stored
noise maps; all rows are paired over the same $N{=}140$ images.}
\label{tab:f1-editfriendly}
\begin{tabular}{lrrrrr}
\toprule
Variant & PSNR & LPIPS & SSIM & sr & $\Delta$ vs.\ M0 ($p$)\\
\midrule
M0\_full (fp32) & 26.86 & 0.065 & 0.778 & 1.000 & ---\\
M1\_fp16 & 26.86 & 0.065 & 0.778 & 0.500 & $+0.00004$ (n.s.)\\
\textbf{M2\_int8} & \textbf{26.73} & 0.066 & 0.777 & \textbf{0.250}
& $\boldsymbol{-0.134}$ ($\boldsymbol{6.4{\times}10^{-23}}$)\\
M3\_int4 & 11.28 & 0.840 & 0.179 & 0.250 & $-15.58$ ($10^{-24}$)\\
M4\_dct\_low (cutoff=8) & 13.39 & 0.808 & 0.514 & 0.016 & $-13.47$ ($10^{-24}$)\\
M5\_random\_proj & 12.92 & 0.805 & 0.493 & 0.125 & $-13.94$ ($10^{-24}$)\\
\bottomrule
\end{tabular}
\end{table}

\subsection{LEDITS++ stored \texorpdfstring{$z_t$}{zt} maps}
\label{app:f1-ledits}

The compressed object is the sequence of stored latent maps used by
LEDITS++ with its native fp16-runtime cache.
\begin{table}[H]\centering\small
\caption{\textbf{F1 transfer to LEDITS++ stored $z_t$ maps.}
\texttt{M0\_full} is the uncompressed native fp16-runtime cache for
this pipeline; the table should be read as a within-method cache
compression comparison.}
\label{tab:f1-ledits}
\begin{tabular}{lrrrr}
\toprule
Variant & PSNR & sr & $\Delta$ vs.\ M0 & $p$\\
\midrule
M0\_full (native fp16) & 15.34 & 1.000 & --- & ---\\
M1\_fp16 cache & 15.34 & 0.500 & $+0.0001$ & 0.72 (n.s.)\\
\textbf{M2\_int8} & \textbf{15.30} & \textbf{0.250} & $\boldsymbol{-0.040}$
& $\boldsymbol{1.0{\times}10^{-2}}$\\
M3\_int4 & 7.23 & 0.250 & $-8.11$ & $10^{-24}$\\
M4\_dct\_low (cutoff=8) & 12.40 & 0.008 & $-2.94$ & $3.2{\times}10^{-23}$\\
M5\_random\_proj & 12.22 & 0.063 & $-3.12$ & $5.1{\times}10^{-23}$\\
\bottomrule
\end{tabular}
\end{table}

\subsection{DDIM-Inv terminal latent}
\label{app:f1-ddim}

The compressed object is the deterministic DDIM inversion terminal
latent $x_T^{\mathrm{inv}}$, which is the transition regime in
Finding~1.
\begin{table}[H]\centering\small
\caption{\textbf{F1 transfer to DDIM-Inv terminal latents.}
\texttt{M0\_full} is the uncompressed native fp16-runtime
$x_T^{\mathrm{inv}}$ cache; this is the transition regime in
Finding~1.}
\label{tab:f1-ddim-inv}
\begin{tabular}{lrrrr}
\toprule
Variant & PSNR & sr & $\Delta$ vs.\ M0 & $p$\\
\midrule
M0\_full (native fp16) & 15.07 & 1.000 & --- & ---\\
M1\_fp16 cache & 15.08 & 0.500 & $+0.0085$ & 0.92 (n.s.)\\
\textbf{M2\_int8} & \textbf{15.07} & 0.250 & $\boldsymbol{-0.0008}$ & $\boldsymbol{0.83}$ (n.s.)\\
M3\_int4 & 14.48 & 0.250 & $-0.585$ & $2.6{\times}10^{-10}$\\
M4\_dct\_low (cutoff=16) & 13.29 & 0.031 & $-1.78$ & $1.9{\times}10^{-17}$\\
M5\_random\_proj & 11.01 & 0.063 & $-4.06$ & $2.5{\times}10^{-24}$\\
\bottomrule
\end{tabular}
\end{table}

\subsection{PnP DirectInv residual cache}
\label{app:f1-pnp}

The compressed object is PnP DirectInv's
\texttt{noise\_loss\_list}; its smaller residual magnitudes place it
in the boundary regime of Finding~1.
\begin{table}[H]\centering\small
\caption{\textbf{F1 transfer to PnP DirectInv residual caches.}
\texttt{M0\_full} is the uncompressed fp32 \texttt{noise\_loss\_list}.
The weaker int8 result places this small-magnitude residual cache in
the boundary regime.}
\label{tab:f1-pnp}
\begin{tabular}{lrrrr}
\toprule
Variant & PSNR & sr & $\Delta$ vs.\ M0 & $p$\\
\midrule
M0\_full (fp32) & 21.20 & 1.000 & --- & ---\\
M1\_fp16 & 21.33 & 0.500 & $+0.13$ & 0.26 (n.s.)\\
\textbf{M2\_int8} & \textbf{20.02} & \textbf{0.250} & $\boldsymbol{-1.18}$ & $\boldsymbol{1.7{\times}10^{-9}}$\\
M3\_int4 & 14.82 & 0.250 & $-6.38$ & $10^{-24}$\\
M4\_dct\_low (cutoff=8) & 16.35 & 0.016 & $-4.85$ & $7.4{\times}10^{-23}$\\
M5\_random\_proj & 15.35 & 0.125 & $-5.85$ & $3.1{\times}10^{-23}$\\
\bottomrule
\end{tabular}
\end{table}

\section{Supplementary tables for F2}
\label{app:f2}

\subsection{P1 trajectory-pair specificity across anchors and backbones}

This table combines the CIFAR $\lambda{=}0.5$ validation check,
conservative SD/SDXL fixed-$\lambda{=}0.7$ references, and the
main \texttt{ramp\_early} $\gamma{=}2$ schedules.
\begin{table}[H]\centering\small
\caption{\textbf{P1 trajectory-pair specificity across models and schedules.}
The forward anchor is compared with random Gaussian, mismatched-image,
shuffled, and sign-flipped anchors under the same reverse setup.
Rows labeled ``main'' use \texttt{ramp\_early} $\gamma{=}2$; fixed
$L_{0.7}$ rows are conservative references.}
\label{tab:p1-anchor-matrix}
\resizebox{\textwidth}{!}{%
\begin{tabular}{llrrrrr}
\toprule
Setup & Anchor & PSNR & LPIPS & SSIM & $\Delta$PSNR vs.\ forward & $p$\\
\midrule
\multirow{6}{*}{CIFAR, $L_{0.5}$, $N{=}2000$}
 & no correction       & 10.40 & 0.196 & 0.116 & $+3.90$ & ---\\
 & \textbf{forward $\epsstar$} & \textbf{14.30} & \textbf{0.165} & \textbf{0.324} & --- & ---\\
 & random Gaussian     & 5.30  & 0.398 & 0.007 & $+9.00$ & $\ll 10^{-22}$\\
 & mismatched-image         & 5.29  & 0.396 & 0.007 & $+9.01$ & $\ll 10^{-22}$\\
 & shuffled            & 5.29  & 0.396 & 0.007 & $+9.01$ & $\ll 10^{-22}$\\
 & sign-flipped        & 5.18  & 0.404 & 0.006 & $+9.12$ & $\ll 10^{-22}$\\
\midrule
\multirow{5}{*}{SD~1.5, $L_{0.7}$, $N{=}140$ (ref.)}
 & forward $\epsstar$ & 17.60 & 0.504 & 0.590 & --- & ---\\
 & random Gaussian     & 6.61  & 0.914 & 0.044 & $+10.98$ & $5{\times}10^{-25}$\\
 & mismatched-image         & 6.60  & 0.915 & 0.044 & $+10.99$ & $5{\times}10^{-25}$\\
 & shuffled            & 6.60  & 0.914 & 0.044 & $+10.99$ & $5{\times}10^{-25}$\\
 & sign-flipped        & 6.52  & 0.918 & 0.042 & $+11.08$ & $5{\times}10^{-25}$\\
\midrule
\multirow{5}{*}{SD~1.5, \texttt{ramp\_early} $\gamma{=}1$, $N{=}140$ (ablation)}
 & forward $\epsstar$ & 23.88 & 0.186 & 0.718 & --- & ---\\
 & random Gaussian     & 6.56  & 0.918 & 0.042 & $+17.32$ & $5{\times}10^{-25}$\\
 & mismatched-image         & 6.55  & 0.919 & 0.042 & $+17.33$ & $5{\times}10^{-25}$\\
 & shuffled            & 6.55  & 0.919 & 0.042 & $+17.32$ & $5{\times}10^{-25}$\\
 & sign-flipped        & 6.50  & 0.921 & 0.040 & $+17.38$ & $5{\times}10^{-25}$\\
\midrule
\multirow{5}{*}{\textbf{SD~1.5, \texttt{ramp\_early}} $\gamma{=}2$, $N{=}140$ (\textbf{main})}
 & \textbf{forward $\epsstar$} & \textbf{24.72} & \textbf{0.147} & \textbf{0.737} & --- & ---\\
 & random Gaussian     & 6.55  & 0.918 & 0.041 & $+18.17$ & $5{\times}10^{-25}$\\
 & mismatched-image         & 6.55  & 0.919 & 0.041 & $+18.18$ & $5{\times}10^{-25}$\\
 & shuffled            & 6.55  & 0.918 & 0.041 & $+18.18$ & $5{\times}10^{-25}$\\
 & sign-flipped        & 6.49  & 0.921 & 0.040 & $+18.23$ & $5{\times}10^{-25}$\\
\midrule
\multirow{5}{*}{SDXL $1024^2$, $L_{0.7}$, $N{=}140$ (ref.)}
 & forward $\epsstar$ & 18.93 & 0.533 & 0.667 & --- & ---\\
 & random Gaussian     & 6.48  & 0.953 & 0.053 & $+12.45$ & $5{\times}10^{-25}$\\
 & mismatched-image         & 6.48  & 0.953 & 0.053 & $+12.45$ & $5{\times}10^{-25}$\\
 & shuffled            & 6.48  & 0.952 & 0.053 & $+12.46$ & $5{\times}10^{-25}$\\
 & sign-flipped        & 6.34  & 0.964 & 0.050 & $+12.59$ & $5{\times}10^{-25}$\\
\midrule
\multirow{5}{*}{SDXL $1024^2$, \texttt{ramp\_early} $\gamma{=}1$, $N{=}140$ (ablation)}
 & forward $\epsstar$ & 25.96 & 0.271 & 0.790 & --- & ---\\
 & random Gaussian     & 6.38  & 0.964 & 0.049 & $+19.58$ & $5{\times}10^{-25}$\\
 & mismatched-image         & 6.38  & 0.964 & 0.049 & $+19.58$ & $5{\times}10^{-25}$\\
 & shuffled            & 6.38  & 0.964 & 0.049 & $+19.58$ & $5{\times}10^{-25}$\\
 & sign-flipped        & 6.29  & 0.971 & 0.047 & $+19.67$ & $5{\times}10^{-25}$\\
\midrule
\multirow{5}{*}{\textbf{SDXL $1024^2$, \texttt{ramp\_early}} $\gamma{=}2$, $N{=}140$ (\textbf{main})}
 & \textbf{forward $\epsstar$} & \textbf{27.12} & \textbf{0.211} & \textbf{0.814} & --- & ---\\
 & random Gaussian     & 6.38  & 0.965 & 0.049 & $+20.74$ & $5{\times}10^{-25}$\\
 & mismatched-image         & 6.38  & 0.965 & 0.049 & $+20.74$ & $5{\times}10^{-25}$\\
 & shuffled            & 6.38  & 0.964 & 0.049 & $+20.74$ & $5{\times}10^{-25}$\\
 & sign-flipped        & 6.29  & 0.971 & 0.047 & $+20.83$ & $5{\times}10^{-25}$\\
\bottomrule
\end{tabular}}
\end{table}

\subsection{P1 after int8 correction-anchor compression}

We additionally include an int8 correction-anchor isolation control
for the main P1 setup. This is a specificity test rather than a
cache-accounted reconstruction: the reverse start is held fixed, and
only the correction anchor is quantized. This checks that compression
does not turn the forward anchor into a generic regularizer.
\begin{table}[H]\centering\small
\caption{\textbf{P1 after int8 correction-anchor compression.}
The start latent is held fixed and only the correction anchor is
quantized; this isolates anchor specificity rather than cache
accounting.}
\label{tab:p1-after-int8}
\begin{tabular}{lrrrrr}
\toprule
Setup & forward/int8 & random & mismatched-image & shuffled & sign-flipped\\
\midrule
SD~1.5, \texttt{ramp\_early} $\gamma{=}2$ & 24.53 & 6.55 & 6.55 & 6.55 & 6.49\\
SDXL, \texttt{ramp\_early} $\gamma{=}2$   & 26.34 & 6.38 & 6.38 & 6.38 & 6.29\\
\bottomrule
\end{tabular}
\end{table}
Thus the compressed anchor still behaves as a trajectory-matched
anchor, not as generic denoising pressure.

\subsection{P4 random-UNet trained-score dependence}

We extend P4 over fixed-$\lambda$ comparators and the two
\texttt{ramp\_early} schedules under the SD~1.5 main reverse setup
($N{=}140$).
\begin{table}[H]\centering\small
\caption{\textbf{P4 trained-score dependence.}
Replacing the SD~1.5 UNet with random weights collapses
reconstruction for every fixed-$\lambda$ comparator and for both
\texttt{ramp\_early} schedules.}
\label{tab:p4-random-unet}
\begin{tabular}{lrrrrr}
\toprule
Schedule / $\lambda$ & V\_trained & V\_random & paired $\Delta$ & $n_{>}/N$ & $p$\\
\midrule
fixed $\lambda{=}0.50$           & 14.82 & 6.92 & $+7.90$  & 140/140 & $5{\times}10^{-25}$\\
fixed $\lambda{=}0.60$           & 15.99 & 7.10 & $+8.89$  & 140/140 & $5{\times}10^{-25}$\\
fixed $\lambda{=}0.70$ (ref.) & 17.60 & 7.42 & $+10.18$ & 140/140 & $5{\times}10^{-25}$\\
fixed $\lambda{=}0.80$           & 19.98 & 8.07 & $+11.91$ & 140/140 & $5{\times}10^{-25}$\\
fixed $\lambda{=}0.90$           & 23.62 & 10.23 & $+13.39$ & 140/140 & $5{\times}10^{-25}$\\
\texttt{ramp\_early} $\gamma{=}1$ (ablation)
                              & 23.88 & 11.66 & $+12.22$ & 140/140 & $5{\times}10^{-25}$\\
\textbf{\texttt{ramp\_early} $\gamma{=}2$ (main)}
                              & \textbf{24.72} & \textbf{12.97} & $\boldsymbol{+11.75}$ & 140/140 & $5{\times}10^{-25}$\\
\bottomrule
\end{tabular}
\end{table}

\subsection{P5 paired statistics for cfg sensitivity}

Rows are image-paired over the $N{=}140$ intersection across all six
fixed-$\lambda$ schedules.

\begin{table}[H]
\centering\small
\caption{\textbf{P5 cfg sensitivity is unimodal in $\lambda$.}
SD~1.5, $N{=}140$ paired across all six $\lambda$ values; main reverse
setup, after-CFG, always blend. Drop $=$ PSNR(cfg=1) $-$ PSNR(cfg=10).}
\label{tab:p5}
\begin{tabular}{lrrrrrrl}
\toprule
$\lambda$ & cfg=1 & 2.5 & 5 & 7.5 & 10 & drop & regime \\
\midrule
0.50           & 16.72 & 16.35 & 15.52 & 14.82 & 14.25 & $2.47$
& binding-insufficient\\
\textbf{0.60}  & 17.95 & 17.58 & 16.75 & 15.99 & 15.36 & $\boldsymbol{2.60}$
& \textbf{plateau}\\
\textbf{0.70} (\textbf{ref.})
               & 19.51 & 19.20 & 18.40 & \textbf{17.60} & 16.91 & $\boldsymbol{2.60}$
& \textbf{plateau (conservative ref.)}\\
0.80           & 21.60 & 21.42 & 20.74 & 19.98 & 19.21 & $2.39$
& post-plateau decay\\
0.90           & 24.63 & 24.55 & 24.17 & 23.62 & 23.00 & $1.63$
& coupling weakening\\
0.95           & 26.38 & 26.35 & 26.18 & 25.90 & 25.57 & $0.81$
& algebraic-collapse limit\\
\bottomrule
\end{tabular}
\end{table}

\begin{table}[H]\centering\small
\caption{\textbf{P5 paired statistics for cfg sensitivity.}
Rows compare the cfg-induced PSNR drop across the fixed-$\lambda$
sweep in Tab.~\ref{tab:p5}.}
\label{tab:p5-paired}
\begin{tabular}{llrl}
\toprule
Comparison & mean $\Delta$ & $p$ & implication\\
\midrule
drop(0.6) vs.\ drop(0.7)   & $-0.005$ dB & $0.916$ & plateau within statistical equivalence\\
drop(0.6) $>$ drop(0.8)    & $+0.20$ dB  & $1.3{\times}10^{-3}$ & plateau upper bound near $0.7$\\
drop(0.8) $>$ drop(0.9)    & $+0.76$ dB  & $7.6{\times}10^{-23}$ & strong monotone decay outside plateau\\
drop(0.7) vs.\ drop(0.95)  & $+1.79$ dB  & $1.22{\times}10^{-24}$ & sensitivity ratio $3.20\times$\\
drop(0.5) vs.\ drop(0.6)   & $-0.12$ dB  & $1.0$ (n.s. opp.) & 0.5 also outside plateau (binding deficit)\\
\bottomrule
\end{tabular}
\end{table}

\section{SDXL transfer: compression, anchor specificity, and editing}
\label{app:sdxl}

\subsection{SDXL reconstruction at \texorpdfstring{\texttt{ramp\_early}}{ramp early}}

This reconstruction transfer uses SDXL at $1024^2$, cfg=7.5,
$N{=}140$, and after-CFG blending.
The schedule transfers cleanly to SDXL.
$\gamma{=}2$ (main): whole-image PSNR $\boldsymbol{27.12}$\,dB /
SSIM $\boldsymbol{0.814}$ / LPIPS $\boldsymbol{0.211}$, paired
$+8.18$\,dB above the conservative fixed $\lambda{=}0.7$ reference
($18.93$ / $0.533$ / $0.667$).
$\gamma{=}1$ (ablation): $25.96$\,dB / $0.790$ / $0.271$, $+7.03$\,dB
over the reference; $\gamma{=}2$ adds $+1.16$\,dB / $-0.060$ LPIPS
on top, matching the SD~1.5 $\gamma$ trend (\S\ref{app:lambda}). The
schedule is written in terms of $\bar\alpha_t$ rather than
UNet-specific internals; the same ${\lambda(\bar\alpha_t)}$ formula
applies to this SDXL backbone.

\subsection{SDXL F1 compression Pareto}

This is the cache-accounted \texttt{ramp\_early} $\gamma{=}2$ SDXL
sweep ($N{=}140$ paired, cfg=7.5).
The cache-accounted SDXL sweep confirms the full-support side of
Finding~1 but also exposes a useful metric split. The int8 anchor is
indistinguishable from the full fp16-runtime anchor, and packed int4
incurs only a small PSNR/LPIPS change. Support-reduced summaries
(DCT-low, random projection, spatial mask, block average) are not
Pareto improvements: they raise PSNR/SSIM through smoothing but
worsen LPIPS by $0.042$--$0.060$ relative to full. The full row here
is the materialized fp16-runtime anchor used by the SDXL pipeline;
storage ratios are normalized to a hypothetical fp32 materialized
anchor, and $\xtstar$ is rebuilt from each decoded payload.
\begin{table}[H]\centering\small
\caption{\textbf{SDXL F1 compression Pareto.}
Cache-accounted \texttt{ramp\_early} $\gamma{=}2$ sweep at
$1024^2$. \texttt{M0\_full} is the fp16-runtime full anchor used by
the SDXL pipeline; storage ratios are normalized to fp32 byte
accounting.}
\label{tab:f1-sdxl}
\begin{tabular}{lrrrr}
\toprule
Scheme & Storage ratio & PSNR & LPIPS & paired $\Delta$PSNR vs.\ M0\\
\midrule
M0\_full (fp16 runtime)       & 0.500 & 27.118 & 0.211 & ---\\
\textbf{M2\_int8 (proposed)} & \textbf{0.250} & \textbf{27.117} & \textbf{0.211}
& $\boldsymbol{-0.0002}$ (n.s.)\\
M3p\_packed int4             & 0.125 & 27.045 & 0.212 & $-0.072$ ($p{=}2.6{\times}10^{-10}$)\\
M4\_DCT low (cutoff=32)      & 0.031 & 27.126 & 0.271 & $+0.008$ (n.s.; LPIPS $+0.060$)\\
M5\_random projection        & 0.063 & 27.431 & 0.262 & $+0.313$ (LPIPS $+0.052$)\\
M7\_spatial mask             & 0.125 & 27.598 & 0.252 & $+0.481$ (LPIPS $+0.042$)\\
M8\_block average            & 0.125 & 27.378 & 0.255 & $+0.260$ (LPIPS $+0.045$)\\
\bottomrule
\end{tabular}
\end{table}

\section{Preliminary SD3/MMDiT Flow-NARC probe}
\label{app:sd3}

\subsection{Mechanism probe summary}

We include this section only as an engineering and scope check. SD~3
uses an MMDiT flow-matching model, so the DDIM $\bar\alpha_t$
correction cannot be copied verbatim. Our pilot uses the
velocity-source Flow-NARC coordinate
$v^{\mathrm{corr}}=(1-\mu_\sigma)\hat v_\theta+
\mu_\sigma(\epsstar-z_0)$ with a non-singular schedule. The $N{=}140$
mechanism probe is suggestive but not main-paper evidence:
anchor specificity is strong, while the trained-vs.\ random
transformer gap is only $+4.87$\,dB / $-0.287$ LPIPS, weaker than the
SD/UNet P4 result.

\subsection{F1 compression pilot on SD~3}

After the mechanism pilot, we ran a small $N{=}40$, 20-step,
$256^2$ compression sweep with the same velocity-source coordinate.
The pattern matches the cautious SDXL reading: element-wise int8 is
indistinguishable from full and packed int4 is close, but
support-reduced summaries trade higher PSNR for much worse LPIPS.
We therefore do not claim DiT F1 transfer in the main paper.
\begin{table}[H]\centering\small
\caption{\textbf{Preliminary SD~3/MMDiT F1 compression pilot.}
Velocity-source Flow-NARC, $N{=}40$, 20 steps, $256^2$ resolution.
\texttt{M0\_full} is the fp16-runtime full flow anchor; this table is
appendix-level preliminary evidence only.}
\label{tab:f1-sd3}
\begin{tabular}{lrrrr}
\toprule
Scheme & Storage ratio & PSNR & LPIPS & Reading\\
\midrule
M0\_full (fp16 runtime)       & 0.500 & 22.550 & 0.169 & reference\\
M1\_fp16                     & 0.500 & 22.550 & 0.169 & same as full\\
\textbf{M2\_int8}            & \textbf{0.250} & \textbf{22.548} & \textbf{0.169} & element-wise stable\\
M3p\_packed int4             & 0.125 & 22.388 & 0.172 & small degradation\\
M4\_DCT low                  & 0.031 & 23.121 & 0.326 & PSNR/LPIPS split\\
M5\_random projection        & 0.063 & 23.538 & 0.312 & PSNR/LPIPS split\\
M7\_spatial mask             & 0.125 & 23.579 & 0.290 & PSNR/LPIPS split\\
M8\_block average            & 0.125 & 23.066 & 0.302 & PSNR/LPIPS split\\
M9\_no anchor                & 0.000 & 8.186  & 0.778 & no correction\\
\bottomrule
\end{tabular}
\end{table}

\section{Editing per-variant tables}
\label{app:edit}

\subsection{SD~1.5 editing variants}

This table uses the mask-rich subset with
\texttt{ramp\_early} $\gamma{=}2$, cfg=7.5, and $N{=}140$.
\begin{table}[H]\centering\small
\caption{\textbf{SD~1.5 editing variants.}
Mask-rich PIE-Bench++ subset, \texttt{ramp\_early} $\gamma{=}2$
int8 main setup, with $\gamma{=}1$ and fixed $\lambda{=}0.7$
mask-background references.}
\label{tab:edit-sd}
\begin{tabular}{lrrrrr}
\toprule
Variant & PSNR & bg LPIPS & CLIP-T & CLIP-dir & NFE\\
\midrule
edit\_no\_narc & 11.83 & 0.413 & 0.275 & 0.117 & 100\\
edit\_whole    & \textbf{25.12} & \textbf{0.087} & 0.244 (worse) & 0.008 (worse) & 100\\
\textbf{edit\_mask\_bg (main)} & 16.96 & 0.104 & 0.265 & 0.100 & 100\\
edit\_mask\_fg (negative control) & 14.28 & 0.397 & 0.265 & 0.060 & 100\\
\midrule
\multicolumn{6}{l}{\emph{Reference rows for comparison:}}\\
edit\_mask\_bg, $\gamma{=}1$ ablation & 16.79 & 0.128 & 0.267 & 0.102 & 100\\
edit\_mask\_bg, fixed $\lambda{=}0.7$ & 15.04 & 0.304 & 0.270 & 0.102 & 100\\
\bottomrule
\end{tabular}
\end{table}

\subsection{SDXL editing variants}

This table repeats the editing comparison at SDXL $1024^2$ with
\texttt{ramp\_early} $\gamma{=}2$, cfg=7.5, and $N{=}140$.
\begin{table}[H]\centering\small
\caption{\textbf{SDXL editing variants.}
PIE-Bench++ subset at $1024^2$, \texttt{ramp\_early} $\gamma{=}2$
int8 main setup, with $\gamma{=}1$ and fixed $\lambda{=}0.7$
mask-background references.}
\label{tab:edit-sdxl}
\begin{tabular}{lrrrrr}
\toprule
Variant & PSNR & bg LPIPS & CLIP-T & CLIP-dir & NFE\\
\midrule
edit\_no\_narc & 12.56 & 0.424 & 0.283 & 0.135 & 100\\
edit\_whole    & \textbf{27.47} & \textbf{0.121} & 0.245 (worse) & 0.011 (worse) & 100\\
\textbf{edit\_mask\_bg (main)} & 17.69 & 0.131 & 0.273 & 0.123 & 100\\
edit\_mask\_fg (neg ctrl) & 15.07 & 0.410 & 0.267 & 0.065 & 100\\
\midrule
\multicolumn{6}{l}{\emph{Reference rows for comparison:}}\\
edit\_mask\_bg, $\gamma{=}1$ ablation & 17.52 & 0.164 & 0.276 & 0.127 & 100\\
edit\_mask\_bg, fixed $\lambda{=}0.7$ & 15.79 & 0.313 & 0.284 & 0.122 & 100\\
\bottomrule
\end{tabular}
\end{table}

\section{Selective gating ablation (fixed-\texorpdfstring{$\lambda$}{lambda} regime)}
\label{app:selective}

\subsection{Fixed-\texorpdfstring{$\lambda$}{lambda} threshold gates}

For the conservative fixed $\lambda{=}0.7$ baseline, we tested
threshold-gated alternatives (hard / soft / boost variants of
$\lambda_t{=}\lambda_{\max}\!\cdot\!\mathbf{1}[r_t{>}\tau_t]$) with
$\tau$ calibrated from the $p_{80}$ percentile of $r_t$ on an
$N{=}140$ pilot. SD~1.5 PSNR/LPIPS/FID results: always blend
$17.60$\,/\,$0.504$\,/\,$146.88$; hard $p_{50}$
$16.16$\,/\,$0.561$\,/\,$151.70$; hard $p_{80}$
$13.97$\,/\,$0.620$\,/\,$151.02$. ``Always blend'' yields the best
of the four on every metric; the CIFAR pattern (selective $<$
always FID) does not transfer to SD latent. A drift-aware ``boost''
variant tuned for high PSNR was found to develop a calibration
failure: under the operational regime $\tau$ is exceeded $\sim
94\%$ of the time, so the variant effectively behaves as a high
fixed-$\lambda$ schedule ($\bar\lambda{\approx}0.94$) with no
Pareto advantage
over its corresponding fixed $\lambda$. The deterministic
\texttt{ramp\_early} schedule (\S\ref{sec:method:schedule}) avoids
this calibration pitfall by being $r_t$-independent.

\section{Mean-\texorpdfstring{$\lambda$}{lambda} matched control}
\label{app:matched-control}

\subsection{Matched-mean fixed schedule}

A natural concern is that the $\gamma{=}2$ improvement over the
conservative $\gamma{=}1$ ablation is merely a higher average anchor
weight ($\bar\lambda=0.885$ vs.\ $0.854$). We therefore run a
matched-mean fixed control with $\lambda_t\equiv0.8852$, keeping the
SD~1.5 setup identical to the main fp16 schedule comparison
($N{=}140$, cfg=7.5, $T{=}50$, after-CFG, same image IDs and
base seed). At identical mean $\lambda$, the ramp remains
substantially better: $+1.73$\,dB PSNR, $+0.019$ SSIM, and
$-0.050$ LPIPS vs.\ the fixed schedule, with paired Wilcoxon
$p\leq5.17{\times}10^{-25}$.

\begin{table}[H]
\centering\small
\caption{\textbf{Mean-$\lambda$ matched control.} Full-anchor SD~1.5
reconstruction, paired by image\_id. The fixed control matches the
$\gamma{=}2$ ramp mean $\bar\lambda$; only the per-step schedule
shape differs.}
\label{tab:matched-control}
\begin{tabular}{lrrrrrr}
\toprule
Schedule & mean $\lambda$ & PSNR & SSIM & LPIPS & paired $\Delta$PSNR & paired $p$\\
\midrule
\texttt{ramp\_early} $\gamma{=}2$ & 0.8852 & \textbf{24.724} & \textbf{0.737} & \textbf{0.147} & \textbf{+1.732} & $5.17{\times}10^{-25}$\\
fixed $\lambda{=}0.8852$          & 0.8852 & 22.992 & 0.717 & 0.197 & --- & ---\\
\bottomrule
\end{tabular}
\end{table}

\section{Hyperparameters and reproducibility}
\label{app:hp}

This appendix records the settings needed to reproduce the reported
numbers. The submitted archive \texttt{NARC-code.zip} provides a README
and \texttt{scripts/run\_reproduction\_suite.sh} with installation
instructions, dry-run-capable paper-scale commands, and the generated-output
layout under \texttt{results/}.

\begin{table}[H]\centering\footnotesize
\caption{\textbf{Main experimental settings.} All SD/SDXL comparisons
are paired by \texttt{image\_id}; SD~3 rows are preliminary appendix
probes and are not used as main evidence.}
\label{tab:hp-settings}
\begin{tabular}{p{0.20\linewidth}p{0.72\linewidth}}
\toprule
Item & Setting \\
\midrule
Cohorts &
PIE-Bench++ \texttt{0\_random\_140} for SD~1.5 and SDXL
($N{=}140$); Null-text is reported on the available $N{=}50$
subset; CIFAR-10 pixel DDPM validation checks use $N{=}2000$; SD~3
uses $N{=}140$ for P1/P4 probes and $N{=}40$ for the F1 compression
pilot. \\
Backbones &
Stable Diffusion~1.5 at $512^2$; SDXL-base at $1024^2$ with the
\texttt{madebyollin/sdxl-vae-fp16-fix} VAE; CIFAR-10
$32{\times}32$ pixel DDPM/DDIM; SD~3 Medium/MMDiT only as a
flow-matching appendix probe. \\
Samplers &
VP/DDPM experiments use deterministic DDIM ($\eta{=}0$) with
$T{=}50$ reverse steps for SD/SDXL and $T{=}100$ for CIFAR. SD~3
uses the Diffusers flow-matching scheduler for $20$ steps at
$256^2$ resolution. \\
Guidance and branch &
Main SD/SDXL reconstruction and editing use cfg $=7.5$ and
after-CFG blending. Mechanism-isolation F1 runs use cfg $=1$ unless
the table states otherwise. SD~3 probes use cfg $=3.0$. \\
Seeds &
Each image uses a deterministic per-image anchor seed
\texttt{derive\_anchor\_seed(image\_id, base\_seed)} with
\texttt{base\_seed=0} in the paper runs. Random-anchor, shuffled,
sign-flipped, mismatched-image, and random-UNet controls use fixed
control seeds recorded in their result directories. \\
\bottomrule
\end{tabular}
\end{table}

\begin{table}[H]\centering\footnotesize
\caption{\textbf{NARC and compression hyperparameters.} The main
operating point is selected by the mechanism tests in
\S\ref{sec:mechanism}; fixed-$\lambda$ rows are retained as
conservative references or ablations.}
\label{tab:hp-narc}
\begin{tabular}{p{0.20\linewidth}p{0.72\linewidth}}
\toprule
Item & Setting \\
\midrule
Main schedule &
\texttt{ramp\_early} with
$(\lambda_{\min},\lambda_{\max},\gamma)=(0.70,0.95,2)$,
mean $\bar\lambda{\approx}0.8852$ for SD~1.5 at $T{=}50$.
The matched-mean fixed control uses $\lambda=0.8852$. \\
NFE accounting &
NFE counts UNet forward calls: \NARC{} $=2T=100$; DDIM-Inv
$=T{+}2T=150$; ReNoise $=T(K{+}1){+}2T$ ($400$ at $K{=}5$);
DCI $\approx112{+}100=212$; Null-text $=609$; EDICT $=8T=400$
with dual streams and CFG split. \\
Reference schedules &
Fixed $\lambda=0.5$, cfg=1 is the mechanism-isolation setup for
the SD F1 compression Pareto and $\lambda$ monotonicity validation
checks. Fixed $\lambda=0.7$, cfg=7.5 is the conservative reference
for Finding~2 analyses and editing ablations. The appendix sweeps
$\lambda\in\{0,0.1,\ldots,1.0\}$ and
$\gamma\in\{0.5,1,2\}$. \\
Cache accounting &
For every compressed-anchor reconstruction row, the stored payload
is decoded once and the same decoded $\tilde\epsilon^\star$ is used
both to rebuild the initial latent $\tilde{x}_T^\star$ and in the
correction term. P1 after-int8 controls are explicitly
correction-anchor isolation tests rather than cache-accounted
reconstructions. \\
Compression schemes &
Full cache, fp16 cast, per-tensor symmetric int8 quantization,
packed signed int4, low-frequency DCT square, regenerable random
projection syndrome, deterministic spatial striding, and block
average. Storage excludes regenerable seeds/projection matrices and
includes materialized tensor payloads plus the recorded scale
metadata. \\
Selective gating &
The main schedule and fixed $\lambda=0.7$ reference use always-blend
and do not use a threshold $\tau$. Selective ablations calibrate
$\tau$ on CIFAR ($N{=}512$, $T{=}100$), SD cfg=1
($N{=}32$, $T{=}50$), or SD cfg=7.5 ($N{=}140$) as stated in the
corresponding appendix table. \\
\bottomrule
\end{tabular}
\end{table}

\subsection{Anchor seed variability audit}
\label{app:seed}

We repeat the main cache-accounted \texttt{ramp\_early} $\gamma{=}2$
int8 row with five deterministic base seeds, changing only
\texttt{base\_seed} in
\texttt{derive\_anchor\_seed(image\_id, base\_seed)}. Seed-level means
are stable: PSNR standard deviation is $0.013$\,dB and LPIPS standard
deviation is $0.00047$ across seeds.

\begin{table}[H]\centering\small
\caption{\textbf{Anchor seed variability audit.}
PIE-Bench++ SD~1.5, $N{=}140$, cfg=7.5, \texttt{ramp\_early}
$\gamma{=}2$, int8 cache-accounted anchor.}
\label{tab:seed-variability}
\begin{tabular}{lrrr}
\toprule
Base seed & PSNR & SSIM & LPIPS\\
\midrule
0 & 24.7241 & 0.7368 & 0.1471\\
1 & 24.7385 & 0.7367 & 0.1470\\
2 & 24.7183 & 0.7365 & 0.1464\\
3 & 24.7022 & 0.7366 & 0.1464\\
4 & 24.7236 & 0.7367 & 0.1474\\
\midrule
Mean $\pm$ sd & $24.7213 \pm 0.0131$ & $0.7366 \pm 0.0001$ & $0.1469 \pm 0.0005$\\
\bottomrule
\end{tabular}
\end{table}

\subsection{Metrics and statistics}

PSNR, SSIM, LPIPS, and MSE are computed per image and then averaged.
All headline comparisons use paired Wilcoxon signed-rank tests by
\texttt{image\_id}; two-sided tests are used for equality and
head-to-head comparisons, and directional alternatives are used only
where a mechanism table states the pre-specified direction. FID is
computed with Inception-v3 features ($d{=}2048$) against the
PIE-Bench++ originals; CLIP-I is reported as an auxiliary
image-similarity check.

\subsection{Compute and software}

All reported SD/SDXL experiments run on a single NVIDIA RTX~4090
(24\,GB). CPU-only unit tests run in under one minute; one SD~1.5
baseline at $N{=}140$ takes roughly $1$--$4$ hours, one SDXL
baseline roughly $1.5$--$3$ hours, one F1 universal sweep roughly
$2$--$4$ hours, and the full Tab.~\ref{tab:main} reproduction about
$24$ hours. Exploratory schedule and ablation development used
additional runs beyond the final reproduction suite, but these
preliminary runs are not required to reproduce the reported tables and
figures; the commands in the supplementary archive target the final
reported experiments. The pinned environment is Python~3.10, PyTorch
2.0.1+cu118, diffusers~0.27.2, and transformers~4.30.2. All wrappers,
analysis scripts, figure-generation scripts, tests, and
reproduction-suite commands needed to regenerate the reported tables are
included in the supplementary archive. Generated CSV/JSON outputs are
written under \texttt{results/} when the reproduction commands are run.
Large cached sample tensors are not included in the archive; commands
that require them regenerate them as intermediate artifacts.

\section{Validation of Lambda Monotonicity and \texorpdfstring{$\gamma$}{gamma} sweep}
\label{app:lambda}

\subsection{11-point fixed-\texorpdfstring{$\lambda$}{lambda} sweep}

At SD~1.5 cfg=1, $N{=}140$, the $11$-point $\lambda$ sweep
$\{0,0.1,\dots,1.0\}$ yields strict monotone PSNR
$12.75\!\to\!27.00$ (each adjacent pair paired $p<10^{-24}$);
$\lambda{=}1$ recovers the SD VAE ceiling (algebraic identity).
$\lambda{=}0.9$ reaches $24.63$\,dB, comparable to
DDIM-Inv@$T{=}50$ at $24.65$\,dB in the same cfg=1 setup while
\emph{halving} the UNet calls (NFE\,$=$\,$50$ vs.\ $100$).

\subsection{\texorpdfstring{$\gamma$}{gamma} sweep at \texorpdfstring{\texttt{ramp\_early}}{ramp early}}

We sweep the schedule's shape parameter
$\gamma\in\{0.5, 1.0, 2.0\}$ keeping
$(\lambda_{\min},\lambda_{\max}){=}(0.70,0.95)$.
$\gamma{=}0.5$: $22.37$\,/\,$0.254$ (PSNR/LPIPS, mean
$\bar\lambda{\approx}0.82$, weakest);
$\gamma{=}1.0$ (ablation): $23.88$\,/\,$0.186$ ($\bar\lambda{\approx}0.854$);
$\gamma{=}2.0$ (main): $24.72$\,/\,$0.147$ ($\bar\lambda{\approx}0.885$).
$\gamma{=}2$ is better on the reported reconstruction and editing
axes; its
trained-vs.-random gap remains large but is slightly smaller than
$\gamma{=}1$ ($+11.75$ vs.\ $+12.22$\,dB), and P5 cfg sensitivity is
weaker but still positive ($+0.69$ vs.\ $+0.88$\,dB). We therefore
use $\gamma{=}2$ as the main quality operating point and retain
$\gamma{=}1$ as the conservative mechanism ablation.

\subsection{\texorpdfstring{$\lambda_{\min}$}{lambda min} sweep at \texorpdfstring{\texttt{ramp\_early}}{ramp early}}

This sweep fixes $\gamma{=}1$, $\lambda_{\max}{=}0.95$, cfg=7.5,
and $N{=}140$.
The low-noise endpoint $\lambda_{\min}$ is intuitively only
``the last few reverse steps'', but on SD's scaled-linear schedule
$\bar\alpha_t$ rises from ${\sim}0$ through ${\sim}0.3$ at
mid-reverse, so $\lambda_{\min}$ in fact governs roughly the
\emph{later half} of the reverse loop. PSNR/LPIPS are monotone
increasing in $\lambda_{\min}$ over $\{0.0, 0.3, 0.5, 0.7\}$:
$19.52$\,/\,$0.337$, $21.16$\,/\,$0.295$, $22.39$\,/\,$0.250$,
$\boldsymbol{23.88}$\,/\,$\boldsymbol{0.186}$ respectively. The
$\lambda_{\min}{=}0.7$ choice is therefore the best-performing tested
operating point under the cfg-plateau-respecting constraint
$\lambda_{\min}{\in}[0.6,0.7]$ (Tab.~\ref{tab:p5}).

\section{Broader impacts and ethical considerations}
\label{app:broader-impacts}

\paragraph{Intended use and positive impacts.}
\NARC{} is intended as a training-free inversion primitive for
reconstruction and retained-source real-image editing studies. By
reducing the additional per-image inversion cache and avoiding
per-image optimization, it may lower the storage and compute burden of
reproducible diffusion inversion experiments and make controlled
reconstruction/editing evaluations more accessible. These benefits are
most relevant to research and service settings where the source image
or clean diffusion state is already retained under an appropriate data
governance policy.

\paragraph{Potential misuse.}
More faithful and cheaper real-image inversion can also lower the
barrier to misleading image edits, unauthorized manipulation of real
images, or other misuse of generative editing systems. \NARC{} does not
introduce a new generative model, scraped dataset, or deployment
system, but it can make existing text-to-image or image-editing
pipelines more efficient in retained-source workflows. We therefore
recommend using it only as part of broader safeguards for generative
editing, such as provenance tracking, watermarking or content
credentials where applicable, access control for editing tools, dataset
and model usage restrictions, and human review for high-risk
deployments.

\paragraph{Privacy and data handling.}
The 16\,KiB figure in this paper refers to additional-cache accounting
in the retained-source editing setting, not standalone archival image
compression. Since retained-source workflows still require access to
the source image, clean diffusion state, or equivalent information,
deployments should apply data minimization, retention limits, access
logging, and deletion policies to both source images and cached
inversion artifacts. Our experiments do not require training on private
user data; they use public benchmarks, pretrained diffusion backbones,
and deterministic per-image seeds for reproducibility.

\section{Existing assets and licenses}
\label{app:assets}

We use existing assets only under their respective licenses and terms
of use. We credit the original papers and providers, name the
version-level models and benchmarks below, and do not redistribute
restricted model weights, benchmark images, or external baseline
repositories. Reproduction instructions direct users to obtain each
asset from its original provider and to follow the corresponding
access conditions. External baseline repositories are cited in the
main text and used under their original licenses; the released archive
contains wrappers and scalar summaries, not third-party repository
snapshots or restricted pretrained weights.

\paragraph{Pretrained models.}
The main SD experiments use
\href{https://huggingface.co/stable-diffusion-v1-5/stable-diffusion-v1-5}
{Stable Diffusion~1.5}~\citep{rombach2022sd}
(CreativeML OpenRAIL-M). The transfer experiments use
\href{https://huggingface.co/stabilityai/stable-diffusion-xl-base-1.0}
{SDXL base~1.0} (CreativeML Open RAIL++-M), together with the
\href{https://huggingface.co/madebyollin/sdxl-vae-fp16-fix}
{SDXL fp16 VAE fix} (MIT). The appendix-level MMDiT probe uses
\href{https://huggingface.co/stabilityai/stable-diffusion-3-medium-diffusers}
{SD~3 Medium}~\citep{esser2024sd3} under the Stability AI
Non-Commercial Research Community License.

\paragraph{Datasets and benchmarks.}
The real-image experiments use
\href{https://huggingface.co/datasets/UB-CVML-Group/PIE_Bench_pp}
{PIE-Bench++ / PIE-Bench}~\citep{ju2023pnp}
(CC-BY-SA-4.0), including the \texttt{0\_random\_140} cohort for the
main reconstruction setting. Pixel-space validation uses CIFAR-10
(Krizhevsky, Nair, and Hinton) through the reference DDPM setup under
the original dataset distribution terms. These datasets are not
redistributed in the supplementary archive.

\paragraph{Metrics and software.}
Software versions are pinned in Appendix~\ref{app:hp}. The software
assets are PyTorch and torchvision (BSD-style), diffusers and
transformers (Apache-2.0), \href{https://github.com/openai/CLIP}{CLIP}
(MIT), and
\href{https://github.com/richzhang/PerceptualSimilarity}{LPIPS}
(BSD-2-Clause).

\clearpage
\section*{NeurIPS Paper Checklist}

\begin{enumerate}

\item {\bf Claims}
    \item[] Question: Do the main claims made in the abstract and introduction accurately reflect the paper's contributions and scope?
    \item[] Answer: \answerYes{}.
    \item[] Justification: The abstract and introduction frame the contribution as two empirical mechanism findings and a training-free \NARC{} operating point, with the main quantitative claims evaluated on PIE-Bench++ SD~1.5 under retained-source additional-cache accounting. The paper also scopes transfer claims to SDXL and states limitations on int8 validation, support-reduced summaries, DiT/SD~3 transfer, and the 16\,KiB storage setting in Section~\ref{sec:discussion}.

\item {\bf Limitations}
    \item[] Question: Does the paper discuss the limitations of the work performed by the authors?
    \item[] Answer: \answerYes{}.
    \item[] Justification: Section~\ref{sec:discussion} includes a dedicated limitations paragraph that bounds the claims around the validated int8 operating point, non-universal behavior of support-reduced summaries, preliminary DiT/SD~3 transfer, retained-source additional-cache accounting, and remaining editing-framework integration. The paper also reports NFE, storage accounting, compute/software details, and seed variability in Appendix~\ref{app:hp}.

\item {\bf Theory assumptions and proofs}
    \item[] Question: For each theoretical result, does the paper provide the full set of assumptions and a complete (and correct) proof?
    \item[] Answer: \answerNA{}.
    \item[] Justification: The paper does not present formal theoretical results such as theorems, lemmas, or proof-based guarantees. Its equations define the forward anchor, correction rule, schedule, and degenerate algebraic endpoint, while the main contributions are empirical mechanism findings and an evaluated training-free inversion primitive.

    \item {\bf Experimental result reproducibility}
    \item[] Question: Does the paper fully disclose all the information needed to reproduce the main experimental results of the paper to the extent that it affects the main claims and/or conclusions of the paper (regardless of whether the code and data are provided or not)?
    \item[] Answer: \answerYes{}.
    \item[] Justification: The paper discloses the information needed to reproduce the main results, including datasets/cohorts, backbones, samplers, guidance settings, operating points, NFE and storage accounting, metrics, paired statistical tests, and seed handling. The submitted \texttt{NARC-code.zip} archive provides the core implementation, experiment wrappers, analysis and figure-generation scripts, unit tests, pinned requirements, a README, and \texttt{scripts/run\_reproduction\_suite.sh}; precomputed \texttt{results/}, \texttt{logs/}, or \texttt{calibration/} directories are not required.

\item {\bf Open access to data and code}
    \item[] Question: Does the paper provide open access to the data and code, with sufficient instructions to faithfully reproduce the main experimental results, as described in supplemental material?
    \item[] Answer: \answerYes{}.
    \item[] Justification: The anonymized supplementary archive \texttt{NARC-code.zip} provides the code needed for the proposed method and experiments, including core \NARC{} primitives, SD/SDXL and CIFAR wrappers, external-baseline wrappers, analysis scripts, figure-generation scripts, tests, environment files, and reproduction-suite commands. The archive intentionally does not include precomputed result files; generated outputs are written under \texttt{results/}, while existing datasets, pretrained models, and external baseline repositories are obtained from their original providers under their access terms.

\item {\bf Experimental setting/details}
    \item[] Question: Does the paper specify all the training and test details (e.g., data splits, hyperparameters, how they were chosen, type of optimizer) necessary to understand the results?
    \item[] Answer: \answerYes{}.
    \item[] Justification: Sections~\ref{sec:method}--\ref{sec:experiments} and Appendix~\ref{app:hp} specify the experimental details needed to interpret and reproduce the results: datasets/cohorts, backbones, samplers, guidance settings, operating points, NFE/storage accounting, metrics, statistical tests, seeds, and \NARC{}/compression hyperparameters. Since \NARC{} is training-free, training optimizer details are not applicable; implementation-level commands are provided in the submitted \texttt{NARC-code.zip} README and reproduction-suite script.

\item {\bf Experiment statistical significance}
    \item[] Question: Does the paper report error bars suitably and correctly defined or other appropriate information about the statistical significance of the experiments?
    \item[] Answer: \answerYes{}.
    \item[] Justification: The paper reports statistical significance for the experiments supporting the main claims, primarily using paired Wilcoxon signed-rank tests by \texttt{image\_id} for compression, mechanism, reconstruction, and editing-related comparisons. Appendix~\ref{app:hp} specifies metric/test computation and reports a five-seed anchor variability audit with mean and standard deviation.

\item {\bf Experiments compute resources}
    \item[] Question: For each experiment, does the paper provide sufficient information on the computer resources (type of compute workers, memory, time of execution) needed to reproduce the experiments?
    \item[] Answer: \answerYes{}.
    \item[] Justification: Appendix~\ref{app:hp} reports the compute and software resources used to reproduce the experiments, including a single NVIDIA RTX~4090 with 24\,GB memory, representative runtimes for SD/SDXL baselines and sweeps, the full main-table reproduction time, and pinned software versions. The appendix also reports NFE accounting for \NARC{} and the evaluated baselines, discloses additional exploratory schedule and ablation runs beyond the final reproduction suite, and states that large cached sample tensors are regenerated as intermediate artifacts by commands that require them.
    
\item {\bf Code of ethics}
    \item[] Question: Does the research conducted in the paper conform, in every respect, with the NeurIPS Code of Ethics \url{https://neurips.cc/public/EthicsGuidelines}?
    \item[] Answer: \answerYes{}.
    \item[] Justification: The authors have reviewed the NeurIPS Code of Ethics, and the research conforms to it. The work is an algorithmic and empirical study using public benchmarks and existing pretrained models, without human-subject experiments, crowdsourcing, private data collection, or special circumstances requiring a deviation from the Code of Ethics.

\item {\bf Broader impacts}
    \item[] Question: Does the paper discuss both potential positive societal impacts and negative societal impacts of the work performed?
    \item[] Answer: \answerYes{}.
    \item[] Justification: Appendix~\ref{app:broader-impacts} discusses positive impacts such as reducing additional per-image inversion cache and compute burden for reproducible diffusion inversion and retained-source editing studies. It also discusses negative impacts such as misleading or unauthorized real-image edits, together with mitigation and data-handling considerations including provenance tracking, watermarking or content credentials, access control, usage restrictions, human review, data minimization, retention limits, access logging, and deletion policies.
    
\item {\bf Safeguards}
    \item[] Question: Does the paper describe safeguards that have been put in place for responsible release of data or models that have a high risk for misuse (e.g., pre-trained language models, image generators, or scraped datasets)?
    \item[] Answer: \answerNA{}.
    \item[] Justification: The paper does not release a new pretrained generative model, high-risk dataset, or scraped data collection requiring controlled-access safeguards. The released materials are code, wrappers, scripts, tests, and documentation for reproducing experiments, while existing pretrained models and benchmarks remain subject to their original access terms.

\item {\bf Licenses for existing assets}
    \item[] Question: Are the creators or original owners of assets (e.g., code, data, models), used in the paper, properly credited and are the license and terms of use explicitly mentioned and properly respected?
    \item[] Answer: \answerYes{}.
    \item[] Justification: Appendix~\ref{app:assets} credits the original providers and papers for the pretrained models, benchmarks, metrics, and software used in the paper, and lists version-level assets, license names, and usage terms where applicable. The README in the anonymized supplementary archive lists the external baseline repositories required by the wrappers. The supplementary archive does not redistribute restricted model weights, benchmark images, or third-party baseline repositories; users are directed to obtain these assets from their original providers.

\item {\bf New assets}
    \item[] Question: Are new assets introduced in the paper well documented and is the documentation provided alongside the assets?
    \item[] Answer: \answerYes{}.
    \item[] Justification: The paper introduces and releases new experimental code rather than a new dataset or pretrained model checkpoint. The submitted \texttt{NARC-code.zip} archive documents these assets with a README, requirements files, core \NARC{} primitives, experiment wrappers, analysis and figure-generation scripts, unit tests, and a reproduction-suite entry point.

\item {\bf Crowdsourcing and research with human subjects}
    \item[] Question: For crowdsourcing experiments and research with human subjects, does the paper include the full text of instructions given to participants and screenshots, if applicable, as well as details about compensation (if any)? 
    \item[] Answer: \answerNA{}.
    \item[] Justification: The paper does not involve crowdsourcing experiments or research with human subjects. The experiments use public benchmarks, existing pretrained diffusion backbones, and deterministic evaluation protocols, so there are no participant instructions, screenshots, or compensation details to report.

\item {\bf Institutional review board (IRB) approvals or equivalent for research with human subjects}
    \item[] Question: Does the paper describe potential risks incurred by study participants, whether such risks were disclosed to the subjects, and whether Institutional Review Board (IRB) approvals (or an equivalent approval/review based on the requirements of your country or institution) were obtained?
    \item[] Answer: \answerNA{}.
    \item[] Justification: The paper does not involve crowdsourcing or research with human subjects, and no participants are recruited, instructed, compensated, or exposed to study-related risks. The experiments use public benchmarks, existing pretrained diffusion backbones, and deterministic evaluation protocols, so there are no participant-risk disclosures or IRB approvals to report.

\item {\bf Declaration of LLM usage}
    \item[] Question: Does the paper describe the usage of LLMs if it is an important, original, or non-standard component of the core methods in this research? Note that if the LLM is used only for writing, editing, or formatting purposes and does \emph{not} impact the core methodology, scientific rigor, or originality of the research, declaration is not required.
    \item[] Answer: \answerNA{}.
    \item[] Justification: LLMs were used only for writing, editing, translation, and formatting assistance and were not used as an important, original, or non-standard component of the core method development. They did not affect the proposed method, experimental protocol, data generation, evaluation, scientific rigor, or originality of the research.

\end{enumerate}

\end{document}